\documentclass[lettersize,journal]{IEEEtran}
\usepackage{amsmath,amsfonts}
\usepackage{array}
\usepackage[caption=false,font=normalsize,labelfont=sf,textfont=sf]{subfig}
\usepackage{textcomp}
\usepackage{stfloats}
\usepackage{url}
\usepackage{verbatim}
\usepackage{graphicx}
\usepackage{subcaption} % 推荐使用的子图宏包

\usepackage{cite}
\usepackage[ruled,vlined]{algorithm2e}
\usepackage{float}

\begin{document}

\title{Biased-Attention Guided Risk Prediction for \\ Safe Decision-Making at Unsignalized Intersections}

\author{Chengyang Dong, Nan Guo}

% \markboth{IEEE TRANSACTIONS ON VEHICULAR TECHNOLOGY,~Vol.~XX, No.~XX, MONTH~202X}%
% {Author \MakeLowercase{\textit{et al.}}: A Deep Reinforcement Learning Approach for...}


\maketitle

\begin{abstract}
Autonomous driving decision-making at unsignalized intersections is highly challenging due to complex dynamic interactions and high conflict risks. To achieve proactive safety control, this paper proposes a deep reinforcement learning (DRL) decision-making framework integrated with a biased attention mechanism. The framework is built upon the Soft Actor-Critic (SAC) algorithm. Its core innovation lies in the use of biased attention to construct a traffic risk predictor. This predictor assesses the long-term risk of collision for a vehicle entering the intersection and transforms this risk into a dense reward signal to guide the SAC agent in making safe and efficient driving decisions. Finally, the simulation results demonstrate that the proposed method effectively improves both traffic efficiency and vehicle safety at the intersection, thereby proving the effectiveness of the intelligent decision-making framework in complex scenarios. The code of our work is available at \url{https://github.com/hank111525/SAC-RWB}.
\end{abstract}


\section{INTRODUCTION}
\IEEEPARstart{W}{ith} the continuous acceleration of urbanization and the latest advancements in autonomous driving technology, Intelligent Transportation Systems (ITS) are undergoing a profound transformation\cite{1,21mokhtari2021pedestriancollisionavoidanceautonomous}. Among these developments, the intelligent management of unsignalized intersections stands as one of the core challenges for enhancing urban traffic efficiency and ensuring driving safety\cite{2}. These intersections are critical nodes within urban road networks; however, due to the lack of explicit right-of-way allocation mechanisms, the complex convergence and weaving of vehicle dynamics can easily lead to traffic congestion and safety conflicts. Therefore, designing a control framework that enables efficient and safe cooperative decision-making for autonomous vehicles in unsignalized intersection scenarios is crucial for realizing next-generation intelligent transportation\cite{3}.

To address the scenario of unsignalized intersections, this issue has been extensively studied and discussed by numerous researchers\cite{23lu2024activead,24chitta,258845649}. Regarding the cooperative control of vehicles at unsignalized intersections, existing research can be broadly categorized into three approaches: rule-based methods, optimization-based methods, and reinforcement learning-based methods.

Early research primarily focused on rule-based methods founded on mathematical models and optimization theory. These methods calculate conflict-free passage plans for each vehicle by precisely planning their trajectories.The author of \cite{32LI2025107960} adopts a robust model predictive control strategy to identify safe gaps and plan trajectories at intersections. The results demonstrate that this algorithm simultaneously improves both traffic efficiency and driving comfort. In literature \cite{4}, the continuous traffic flow is discretized into a series of small-scale vehicle groups, and a cooperative grouping-based control method is proposed for unsignalized intersections. This method decomposes vehicle motion into two processes: branch adjustment and intersection coordination. Concurrently, a Cooperative Adaptive Cruise Control (CACC) model with branching is introduced for vehicle tracking control. Simulation results demonstrate that the proposed method can achieve smoother vehicle operation while reducing energy consumption by up to 29\%. Literature \cite{5} proposed a cooperative motion planning method that integrates offline optimization learning with online decision-making. First, a Learning-Based Iterative Optimization (LBIO) algorithm is used offline to generate speed-optimal, collision-free spatio-temporal trajectories for all potential vehicle conflict patterns. 
During the online application phase, a Monte Carlo Tree Search (MCTS) algorithm is utilized to determine the optimal vehicle passing sequence based on real-time traffic flow, and vehicles are dynamically partitioned into distinct "clusters" to ensure safe and efficient intersection traversal.
The authors of \cite{3010.1007/s10489-011-0322-z} developed an optimization model to optimize vehicle passing sequence and travel time, aiming to minimize the service time for all vehicles. Reference \cite{31AHMANE201344} adopted the waiting time of all vehicles as the loss function. However, with the increase in the number of vehicles, such methods face challenges in multi-participant scenarios due to the curse of dimensionality and computational burden \cite{29CHEN2022127953}.

Reinforcement Learning (RL), as a decision-making algorithm, enables the learning of effective policies in complex environments \cite{6bouton2019reinforcement,269294407}.As the complexity of the problem increases, Deep Reinforcement Learning (DRL) methods have garnered widespread attention for their powerful autonomous learning and decision-making capabilities \cite{19pattanayak2024deep}. Researchers have leveraged DRL to address the uncertainty and high-dimensional state spaces inherent in intersection scenarios. For instance, in literature \cite{8spatharis2024multiagent}, routes rather than vehicles are defined as agents, and multi-agent cooperation is achieved by introducing a predictive collision term into the state. The authors validated the effectiveness of this approach through experiments in various scenarios.
Some studies have focused on stochastic task scenarios. To tackle the high uncertainty challenge arising from random driving tasks (e.g., going straight, turning left, turning right) at unsignalized intersections, the literature \cite{9xiao2024decision} designed a Mix-Attention network to filter critical environmental information. Furthermore, task variables were incorporated into the state to distinguish driving intentions. Simulation results confirmed that the model achieved superior comprehensive performance in both safety and traffic efficiency compared to the original baseline.
To enhance safety, some researchers have explored how to integrate safety constraints into the DRL framework to achieve more reliable driving decisions. Literature \cite{7yang2024towards} proposed a hierarchical safety decision-making framework that quantifies the epistemic uncertainty of decisions online to assess the reliability of the primary RL policy in real-time. Simulation results demonstrated that this framework could effectively handle unseen scenarios during training, thereby reducing the risk of collisions.
Reference \cite{34yu2025uncertainty} developed an Uncertainty-aware Safety-critical Decision and Control (USDC) framework, which generates risk-averse policies by constructing a risk-aware inheritable distributional reinforcement learning model.
Recently, Wang et al. \cite{10wang2025learning} proposed a method to guide and coordinate mixed traffic flow using a small number of autonomous Robotic Vehicles (RVs). This method makes decisions in a high-level discrete action space (go/stop) and relies on an external conflict resolution mechanism to ensure safety. Additionally, to make the behavior of autonomous vehicles more understandable and acceptable to human drivers, other studies have utilized reinforcement learning to train vehicles to exhibit “polite” driving behaviors \cite{11yan2021courteous}.

However, vehicle control strategies based on Deep Reinforcement Learning still face the following challenges: 1) Sparse and Catastrophic Collision Penalties: Many existing reinforcement learning methods primarily rely on sparse, catastrophic collision penalties (e.g., assigning a large negative reward upon a collision) to learn safe policies. The agent must learn to avoid hazards through extensive trial-and-error, which results in low learning efficiency and often leads to overly conservative driving strategies that converge. 2) Lack of Long-Term Risk Assessment: Traditional RL reward functions often focus on instantaneous states, such as the current speed or the distance to the preceding vehicle, while lacking an assessment of future long-term risks. For instance, an incorrect decision made at the current time step might not cause a collision, but a subsequent reasonable decision could lead to one. This indicates a deficiency in proactive risk perception capabilities. 3) Imbalanced Experience from Collision Events: During the initial stages of training, collision events occur frequently. This leads to an imbalance in the experience pool, where collision-related experiences are either overrepresented or underrepresented compared to safe driving events. Such an imbalance hinders the model’s ability to be adequately trained, thereby compromising its final performance and stability.

To transcend the conventional “reactive-safety” paradigm and realize a proactive-safety regime for unsignalized intersection negotiation, this paper proposes a biased-attention-augmented deep reinforcement-learning (DRL) decision framework built upon the Soft Actor-Critic (SAC) algorithm. The methodology comprises two synergistic contributions. First, a transformer-based risk predictor is devised that maps the historical state-decision sequence into a continuous risk coefficient; this coefficient is subsequently embedded into the SAC reward function as a dense, risk-aware reward, thereby mitigating the notorious sparse-reward problem and dissolving the myopic bias toward instantaneous states. Second, a hierarchical experience buffer is constructed in which heterogeneous transitions are stored in dedicated sub-buffers. This stratified replay mechanism accelerates convergence by providing the learner with contextually coherent mini-batches that respect the intrinsic temporal structure of intersection dynamics.

The main contributions of this paper are summarized as follows:

We propose a Soft Actor-Critic (SAC) decision-making framework integrated with a biased attention mechanism. This framework combines the Transformer model with the SAC reinforcement learning algorithm to address the coordinated control task of continuous speed and passing order at intersection scenarios.
We leverage the sequential modeling capability of the Transformer to construct a model that can learn and predict long-term collision risks from vehicle historical trajectories and the current traffic situation. This model transforms the predicted risk into a dense reward signal, providing the RL agent with forward-looking safety guidance to proactively avoid future potential dangers.

The remainder of this paper is structured as follows. First, Section \ref{PROBLEM FORMULATION} defines the scenario and vehicles, formulates the problem as a Markov Decision Process (MDP), and details the design of the state, action, and reward. Section \ref{METHODOLOGY} presents the overall framework of the proposed method, including the Transformer-based risk predictor and the hierarchical experience replay mechanism. Section \ref{EXPERIMENTS} provides the experimental setup and an analysis of the simulation results. Finally, Section 5 offers the conclusion and outlines future work.

\section{PROBLEM FORMULATION}
\label{PROBLEM FORMULATION}
\begin{figure}[ht]
    \centering
    \includegraphics[width=0.5\textwidth]{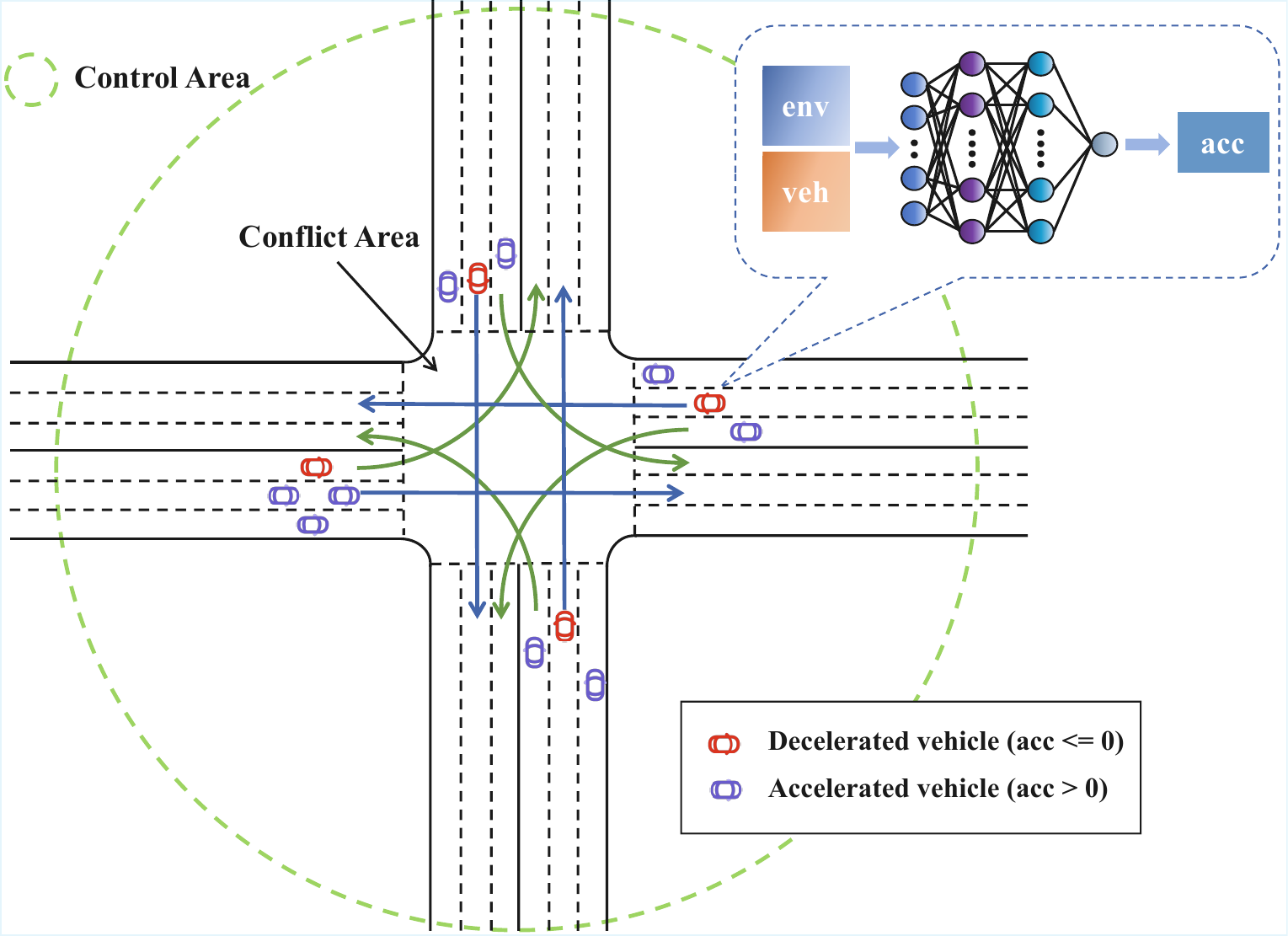}
    \caption{Schematic Diagram of the Intersection Scenario}
    \label{fig:1}
\end{figure}
In this section, we provide a detailed description of the unsignalized intersection scenario and the modeling approach for environmental vehicles, framing the vehicle decision-making problem as a multi-agent reinforcement learning (MARL) task. The discussion begins with the construction of a four-way intersection environment. This is followed by an explanation of the low-level controller of the vehicle. The section concludes with a presentation of the formulation of the Markov Decision Process (MDP) model.

\subsection{Scenario Definition}

\begin{figure*}[t]
    \centering
    \includegraphics[width=\textwidth]{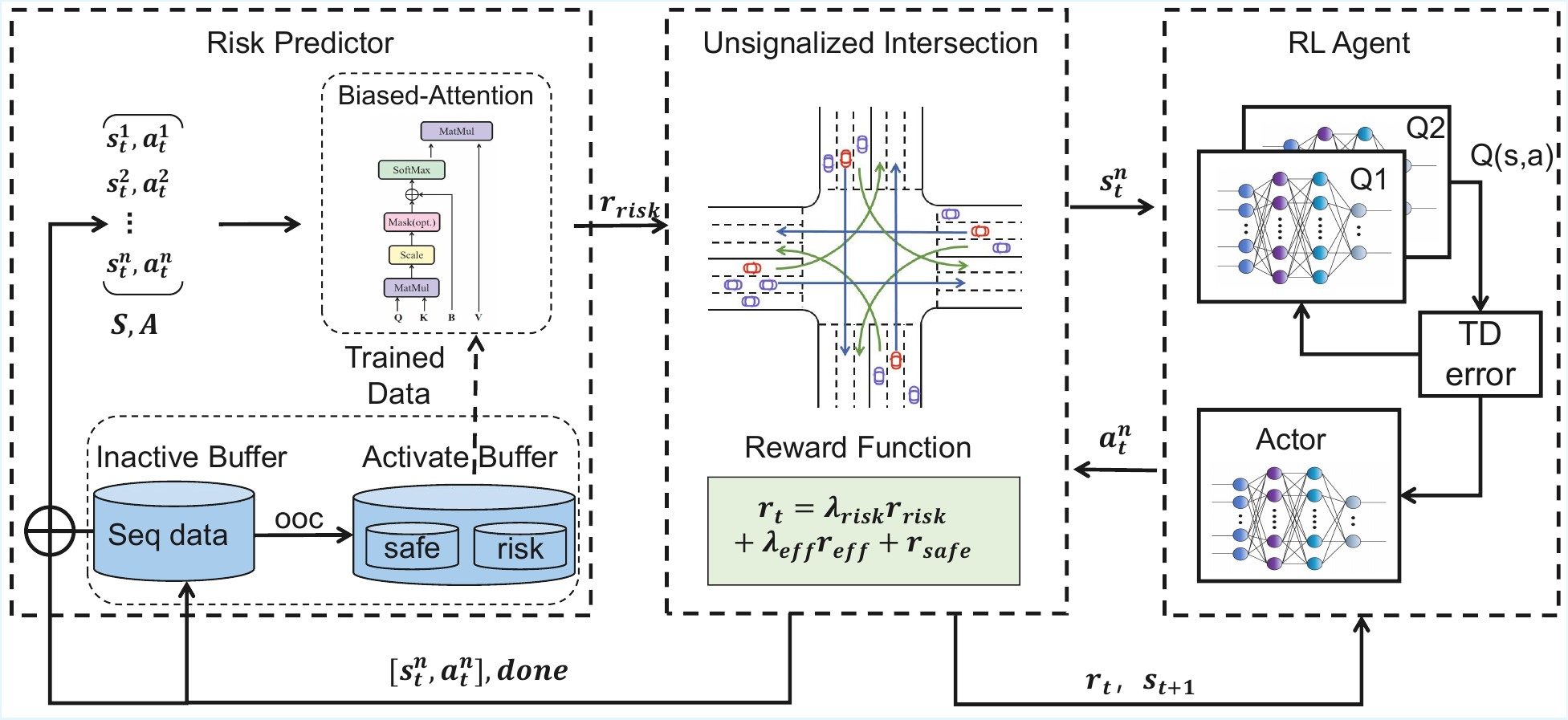}
    \caption{Schematic of the decision framework. \textbf{Note:} OOC stands for Out of Control (vehicle has lost control).}
    \label{fig:framework}
\end{figure*}

We consider a classic unsignalized intersection scenario, which consists of four roads, each with $K \geq 3$ lanes, as illustrated in Figure~\ref{fig:1}. 
Each lane is designated for a single specific task, such as a left turn. 
This setting effectively distinguishes the passing intentions of vehicles, prevents random maneuvers, and enhances driving safety. 

The blue region in the figure denotes the control zone. 
The objective of this paper is to control the vehicles entering the intersection, guiding them to enter the intersection at an appropriate speed to ensure a safe and orderly passing sequence. 
The core goal of our research is to investigate the longitudinal behavioral decision-making of autonomous vehicles.

Specifically, when a vehicle's distance to the intersection is less than a given threshold (set to 30\,m in this paper), it is designated as a \textbf{controlled vehicle} (the dark blue vehicle in \ref{fig:1}). Through a series of control actions, this vehicle is guided to traverse the intersection safely.

For a single intersection, vehicles approaching from the four cardinal directions---East, West, South, and North---are denoted as $E$, $W$, $S$, and $N$, respectively. At the intersection, vehicles have three possible maneuvers: \textbf{go left (GL)}, \textbf{go straight (GS)}, and \textbf{go right (GR)}. In general scenarios, most intersections are typically equipped with right-turn lanes, and relevant studies have focused on vehicle right-turn scenarios \cite{18li2024decision,2710537094,32LI2025107960}. Therefore, this study does not consider right-turn maneuvers. Under this assumption, there are eight possible movement directions for a single intersection: $(E-\text{GL})$, $(E-\text{GS})$, $(W-\text{GL})$, $(W-\text{GS})$, $(N-\text{GL})$, $(N-\text{GS})$, $(S-\text{GL})$, and $(S-\text{GS})$.

\subsection{Vehicle Control Model}
The Intelligent Driver Model (IDM) \cite{12treiber2000congested} is a microscopic traffic flow model that simulates longitudinal driving behavior, specifically car-following. It is founded on a set of well-defined physical and behavioral assumptions, employing mathematical formulations to characterize how a driver smoothly modulates their acceleration based on the state (i.e., speed and distance) of the front vehicle.

Compared to other models, the IDM exhibits greater applicability. Firstly, the parameters of the IDM, such as the desired speed and safe time headway, have clear physical meanings. This makes it more intuitive to understand and calibrate than complex psycho-physical models like the Wiedemann model. Secondly, the model generates continuous and smooth acceleration trajectories, which more realistically reflect vehicle dynamics compared to discrete models such as Cellular Automata. Furthermore, the IDM theoretically guarantees collision-free behavior, providing a reliable level of safety for simulations \cite{16zhang2024car}.

In this paper, vehicles in the non-controlled area are managed using the Intelligent Driver Model (IDM). The main parameter settings for the IDM are listed in Table \ref{tab:IDM}.

% Please ensure you have \usepackage{float} in your preamble
\begin{table}[H]
    \centering
    \caption{IDM Parameters}
    \label{tab:IDM}
    \begin{tabular}{lccc}
        \hline
        \textbf{Parameter} & \textbf{Value} & \textbf{Unit} \\
        \hline
        Maximum acceleration & 3 & $m/s^2$ \\
        Target time interval & 1.5 & s \\
        Maximum deceleration & -5 & $m/s^2$ \\
        Desired jam distance & 10 & m \\
        \hline
    \end{tabular}
\end{table}

\subsection{Markov Decision Process}
To solve the autonomous driving problem at unsignalized intersections using Reinforcement Learning (RL), we formulate this problem as a Markov Decision Process (MDP). An MDP provides a standard mathematical framework for an agent to learn an optimal policy through interaction with its environment. Our goal is to find an optimal policy, denoted as $\pi^*$, which maps environmental states to the agent's actions, thereby maximizing the long-term cumulative reward. This MDP is defined by a tuple $(S, A, R, P, \gamma)$, whose specific components are described as follows.

\subsubsection{State Space}
The state space $\mathcal{S}$ is a complete description of the environment in which the agent is situated at a given time, and it serves as the foundation for the agent’s decision-making. In this study, the state space is composed of the agent’s own information and its perceptual information about the intersection environment.

The state space $\mathcal{S}$ is defined by a set of features representing the vehicle's condition and its surrounding traffic context. A state at any time $t$, denoted as $s_t \in \mathcal{S}$, is composed of the following components:
\begin{equation}
    s_t = \{d_{\text{inter}}, v_{\text{control}}, U_S, U_Q, U_G\},
\end{equation}
where the components are defined as follows:
\begin{itemize}
    \item \textbf{$d_{\text{inter}}$}: Represents the distance from the front of the vehicle to the intersection stop line. This value is normalized by dividing by the control distance of 30 m, resulting in a range of $d_{\text{inter}} \in [0, 1]$.

    \item \textbf{$v_{\text{control}}$}: Represents the normalized speed of the vehicle. It is obtained by dividing the current vehicle speed by the maximum speed, with a value range of $v_{\text{control}} \in [0, 1]$.

    \item \textbf{$U_S$}: The set of waiting times, $U_S = \{u'_{s1}, \dots, u'_{s8}\}$. Each element is normalized by dividing by a preset maximum waiting time $T_{\max}$ (where $T_{\max} = 300$), i.e., $u'_{si} = u_{si} / T_{\max}$. The normalized values are in the range $u'_{si} \in [0, 1]$.

    \item \textbf{$U_Q$}: The set of queue lengths, $U_Q = \{u'_{q1}, \dots, u'_{q8}\}$. Each element is normalized by dividing by the maximum number of vehicles in a lane, $Q_{\max}$, i.e., $u'_{qi} = u_{qi} / Q_{\max}$. The normalized values are in the range $u'_{qi} \in [0, 1]$.

    \item \textbf{$U_G$}: Represents the occupancy state of the intersection area, used to describe internal congestion. For each approach lane, the internal part of the intersection is divided into 10 equal-sized blocks. This state component consists of the vehicle status in these blocks for all eight directions. If a block is occupied, its value is the normalized speed of the vehicle; otherwise, it is 0. Thus, $U_G = \{u'_{g1}, \dots, u'_{g80}\}$, with values in the range $u'_{gi} \in [0, 1]$.
\end{itemize}

\subsubsection{Action Space}

The objective of reinforcement learning is to map an observed state to a control command for the controlled vehicle. Since this paper focuses on controlling the vehicle's behavioral decisions, the action space $\mathcal{A}$ is defined as the vehicle's acceleration:
\begin{equation}
    \mathcal{A} = \{acc\} \quad \text{s.t.} \quad acc \in [-3, 3] \, m/s^2
    \label{eq:action_space}
\end{equation}

To facilitate the training process, the agent's network outputs a value in the range of $[-1, 1]$, which is then linearly scaled to the defined acceleration space.

\subsubsection{Reward Function} % 3) Reward Function

To encourage the controlled vehicle to simultaneously consider its own driving efficiency and the risk of conflicts within the intersection, we designed a multi-component reward function. It incorporates a risk-aware reward and an efficiency reward to guide the agent toward making globally optimal decisions. The total reward is composed of the following parts:

\begin{itemize}
    \item \textbf{Driving Efficiency Reward):} 
    This component incentivizes efficient passage through the intersection. It is calculated based on the vehicle's acceleration and its waiting time rank relative to other vehicles. The formula is as follows:
    \begin{equation}
        r_{\text{eff}} = \left( \text{rank}_i - \frac{1}{8} \sum_{j=1}^{8} \text{rank}_j \right) \cdot \frac{acc_i}{acc_{max}},
    \end{equation}
    where $\text{rank}_i$ is the rank of vehicle $i$ based on its waiting time, and $acc_i$ is its current acceleration.
    
    \item \textbf{Collision Penalty:} 
    A large negative reward is given if the vehicle is involved in a collision to discourage unsafe actions strongly.
    \begin{equation}
        r_{\text{safe}} = 
        \begin{cases} 
            -acc_i/acc_{max} & \text{if conflict} \\
            acc_i/acc_{max}   & \text{otherwise}
        \end{cases}
    \end{equation}
    
    \item \textbf{Risk Reward:} 
    This reward is derived from the output of the risk predictor module. It provides a dense, forward-looking signal about the long-term safety of the current policy.The output layer of the risk predictor module uses the sigmoid function to get $r_{\text{risk}} \in [0, 1]$. This component is discussed in detail in Section \ref{Risk Predictor Based on Biased Attention}.
\end{itemize}

The total reward $r$ is a weighted sum of these components:
\begin{equation}
    r = \lambda_{\text{eff}} r_{\text{eff}} + \lambda_{\text{risk}} r_{\text{risk}} + \lambda_{\text{safe}} r_{\text{safe}}
    \label{eq:total_reward},
\end{equation}
where $\lambda_{\text{eff}}$, $\lambda_{\text{risk}}$ and $\lambda_{\text{safe}}$ are weighting coefficients for the efficiency, risk and safe reward, respectively.

\subsection{Optimization Objective}

For complex decision-making tasks such as those at intersections, a policy that solely maximizes the cumulative reward is prone to settling in local optima and may lack adaptability to unseen situations. To learn a more exploratory and robust driving policy, we introduce \textbf{Policy Entropy} into the optimization objective, framing the problem within the maximum entropy reinforcement learning framework. The entropy $\mathcal{H}$ of a policy $\pi$ in a given state $s_t$ is defined as:
\begin{equation}
    \mathcal{H}(\pi(\cdot|s_t)) = \mathbb{E}_{a_t \sim \pi(\cdot|s_t)}[-\log\pi(a_t|s_t)]
\end{equation}

The optimization objective is thus to maximize the \textbf{entropy-regularized} cumulative return. The optimal policy $\pi^*$ must not only maximize the reward but also maintain as much randomness as possible:
\begin{equation}
    \pi^{*} = \arg\max_{\pi} \mathbb{E}_{\tau \sim \pi} \left[ \sum_{t=0}^{T} \gamma^t \left( r(s_t, a_t) + \alpha \mathcal{H}(\pi(\cdot|s_t)) \right) \right]
    \label{eq:max_entropy_objective}
\end{equation}
where $\tau = (s_0, a_0, s_1, a_1, \dots)$ is a trajectory generated by the policy $\pi$. The parameter $\alpha > 0$ is the \textbf{temperature coefficient}, which balances the trade-off between reward maximization (exploitation) and entropy maximization (exploration). This objective drives the agent to learn a policy that is both efficient and highly exploratory, ultimately enabling safe and effective navigation through unsignalized intersections.

\section{METHODOLOGY}
\label{METHODOLOGY}
\subsection{Overall Framework}
This paper proposes a SAC decision-making framework that integrates risk perception and efficient learning. This framework is an end-to-end perception-decision-learning closed-loop system, enabling autonomous vehicles to make proactive and safe driving decisions in complex unsignalized intersection environments.

The architecture of the proposed framework is illustrated in Figure~\ref{fig:framework} . It is primarily composed of three components: a risk predictor, an intersection simulation environment, and an agent. First, the environment provides a corresponding intersection scenario for each vehicle based on the current operational state of the intersection. Subsequently, the agent employs the SAC algorithm to generate an intelligent driving policy based on the environmental state. The environment then computes an efficiency reward and a safety reward according to this policy. Furthermore, based on historical states and decision-making information, the environment invokes the risk predictor to assess the risk of the current decision, converting this risk into a risk reward. These rewards are combined to form a total reward that accounts for both efficiency and safety. Finally, the SAC and Transformer models sample from the experience replay buffer to update their networks, continuously optimizing the driving policy.

In the following subsections, we will introduce each core component in detail: the foundational decision-making algorithm, SAC (Section \ref{SAC Algorithm}); the Transformer-based safety risk predictor (Section \ref{Risk Predictor Based on Biased Attention}); and the hierarchical experience replay mechanism (Section \ref{Hierarchical Experience Replay Mechanism}).

\subsection{SAC Algorithm}
\label{SAC Algorithm}
To solve the Markov Decision Process (MDP) defined in \ref{PROBLEM FORMULATION}, we employ the Soft Actor-Critic (SAC) \cite{13haarnoja2018soft} algorithm as the core decision-making module. SAC is an off-policy Deep Reinforcement Learning (DRL) algorithm specifically designed for tasks with continuous action spaces.

The SAC algorithm is grounded in the maximum entropy reinforcement learning framework. Unlike traditional reinforcement learning, which solely aims to maximize the cumulative reward, the maximum entropy framework seeks to maximize both the cumulative reward and the entropy of the policy. By introducing entropy as a regularization term, this approach encourages the agent to explore more thoroughly, prevents the policy from prematurely converging to a local optimum, and consequently enhances the policy’s robustness.

Soft policy iteration primarily involves two stages: policy evaluation and policy improvement. The SAC algorithm consists of two types of neural networks: an Actor network for policy optimization and a Critic network for policy evaluation.

The \textbf{Critic network}'s task is to evaluate the policy accurately. It is updated by minimizing the Mean Squared Error (MSE) between its predicted Q-value and a more accurate "target Q-value" denoted by $y$. This target value is a concrete representation of the maximum entropy Bellman equation:
\begin{equation}
    y = r(s, a) + \gamma \left( \min_{i=1,2} Q_{\theta'_i}(s', a') - \alpha \log \pi_{\phi}(a'|s') \right),
\end{equation}
where $a' \sim \pi_{\phi}(\cdot|s')$
 
The loss function for the Critic network, $J_Q(\theta_i)$, is defined as follows:
\begin{equation}
    J_Q(\theta_i) = \mathbb{E}_{(s,a,r,s') \sim \mathcal{D}} \left[ \frac{1}{2} (Q_{\theta_i}(s, a) - y)^2 \right],
\end{equation}
where $\mathcal{D}$ represents the experience replay buffer. Minimizing this loss enables the Critic network to learn a more accurate value assessment of the current policy.

The \textbf{Actor network}'s task is to optimize its policy based on the evaluation from the Critic network. Its objective is to adjust the policy to produce actions that yield higher Q-values and greater entropy. The Actor's loss function $J_{\pi}(\phi)$ is given by:
\begin{equation}
    J_{\pi}(\phi) = \mathbb{E}_{s \sim \mathcal{D}, a \sim \pi_{\phi}} \left[ \alpha \log(\pi_{\phi}(a|s)) - \min_{i=1,2} Q_{\theta_i}(s, a) \right]
\end{equation}
By minimizing this loss, the policy is updated in a direction that maximizes the entropy-regularized return.

\begin{figure}[ht]
    \centering
    \includegraphics[width=0.4\textwidth, height=8cm, keepaspectratio]{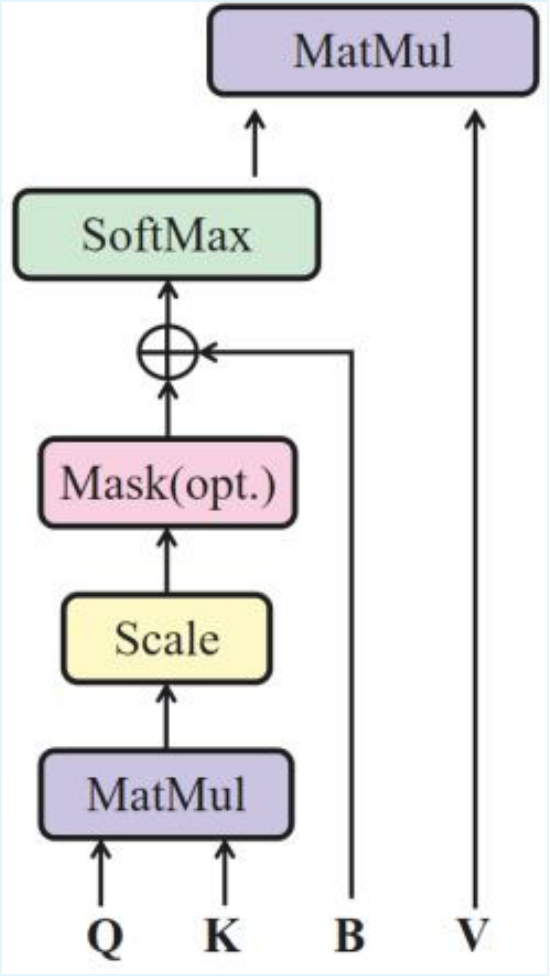}
    \caption{Biased Attention Structure}
    \label{fig:3}
\end{figure}

\subsection{Risk Predictor Based on Biased Attention}
\label{Risk Predictor Based on Biased Attention}
In 2017, the Transformer model based on the self-attention mechanism was proposed in \cite{14vaswani2017attention}. Subsequently, the attention mechanism has been widely applied across various fields. Its foundation is the Scaled Dot-Product Attention, which calculates the similarity scores between a Query and all Keys to assign corresponding weights to the Values.

To enhance the model’s expressive power, the Transformer further employs the Multi-Head Attention mechanism. This mechanism operates multiple independent attention heads in parallel and concatenates their outputs, allowing the model to jointly attend to information from different representation subspaces at various positions. Since the self-attention mechanism itself is inherently unaware of sequence order, Positional Encoding is also introduced. It injects absolute or relative positional information into the input vectors to preserve sequential characteristics.

\begin{figure*}[ht]
    \centering
    \includegraphics[width=\textwidth]{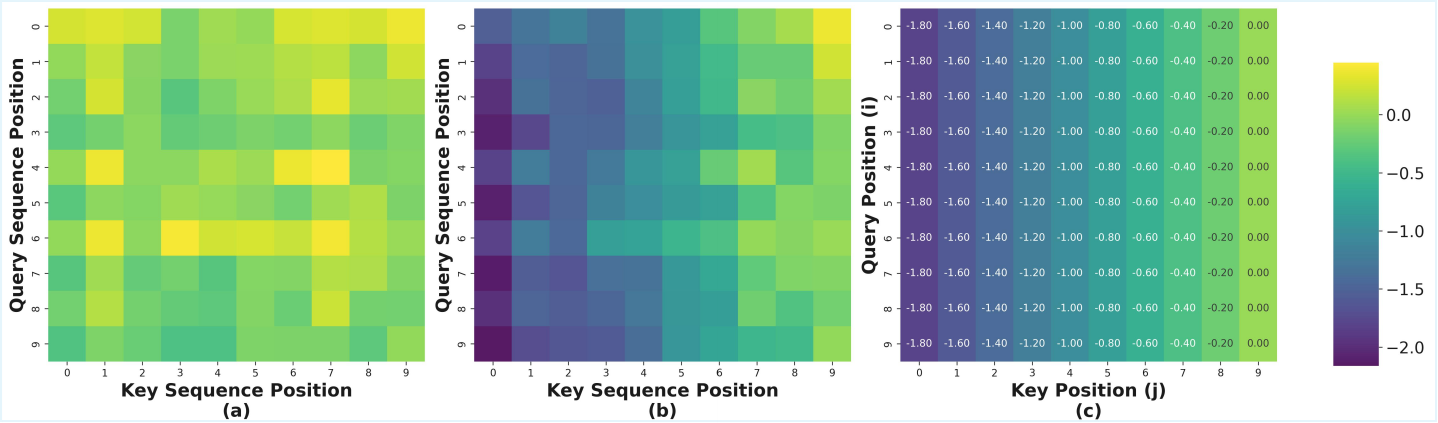}
    \caption{Biased Attention Visualization. (a) Raw Attention Scores ($QK^T/\sqrt{d_k}$). (b) Scores with Bias ($QK^T/\sqrt{d_k} + B$). (c) Attention Bias Matrix.}
    \label{fig:attention_visualization}
\end{figure*}

Risk assessment in unsignalized intersection scenarios is highly dependent on the behavioral sequences of vehicles over time. The self-attention mechanism, with its ability to effectively capture long-range dependencies in sequential data, is naturally suited for modeling and predicting traffic trajectories. Reference \cite{17press2021train} proposed Attention with Linear Biases (ALiBi), which achieves efficient extrapolation by introducing distance-proportional linear biases into attention scores. Considering that intersection decision-making scenarios involve strong prior knowledge, where events occurring closer in time are more critical for risk prediction, we construct an attention bias matrix, denoted as B, within the self-attention mechanism, as illustrated in Figure~\ref{fig:3}. This matrix applies a positive additive bias to later time steps during standard attention score computation, thereby encouraging the model to prioritize recent states and actions during the learning process.

The model's risk prediction process can be described as follows. First, we concatenate the state-action pairs $(s_i, a_i)$ from the most recent $n$ timesteps into an input sequence $X = [x_1, \dots, x_n]$, where $x_i = [s_i; a_i]$. This sequence is then passed through an embedding layer and positional encoding, followed by a linear mapping to generate the query matrix $Q$, key matrix $K$, and value matrix $V$. We then modify the standard attention score with the bias matrix $B$. The final formula is:
\begin{equation}
    \text{Attention}(Q, K, V) = \text{softmax}\left(\frac{QK^T}{\sqrt{d_k}} + B\right)V
\end{equation}
The bias matrix $B$ is an $N \times N$ matrix, where each element $B_{i,j}$ is defined as:
\begin{equation}
    B_{i,j} = \beta \cdot (j - (N-1)),
\end{equation}
where $i$ and $j$ represent the indices of the query and key positions in the sequence ($i, j \in \{0, 1, \dots, N-1\}$), $N$ is the input sequence length (set to 10), and $\beta$ is a decay factor that controls the strength of the bias (set to 0.2). Figure~\ref{fig:3} illustrates the structure with the added bias, and Figure~\ref{fig:attention_visualization} provides a visualization. Figure~\ref{fig:attention_visualization}a shows the raw attention matrix, Figure~\ref{fig:attention_visualization}c is the bias matrix, and Figure~\ref{fig:attention_visualization}b is the attention matrix after adding the bias. The brighter the color, the higher the attention score. As shown in Figure~\ref{fig:attention_visualization}b, the attention scores are highest towards the right side (representing the most recent decisions), indicating that the bias matrix effectively assigns the highest attention to the latest decisions.

The classifier is a multi-layer perceptron (MLP) with three layers, containing [128, 64, 1] neurons respectively, using ReLU as the activation function for each hidden layer. Finally, a Sigmoid activation function is applied to produce a predicted risk value between [0, 1]. The computational process is as follows:

\begin{align}
    z_0 &= X + E_{\text{pos}}, & E_{\text{pos}} \in \mathbb{R}^{N \times D} \\
    z'_{l} &= \text{MSA}(\text{LN}(z_{l-1})) + z_{l-1}, & l=1 \dots L \\
    z_{l} &= \text{MLP}(\text{LN}(z'_{l})) + z'_{l}, & l=1 \dots L \\
    h_1 &= \text{ReLU}(W_1 z_L^N + b_1) \\
    h_2 &= \text{ReLU}(W_2 h_1 + b_2) \\
    r_{\text{risk}} &= \text{Sigmoid}(W_3 h_2 + b_3)
\end{align}
% \begin{align}
%     \mathbf{x} &= \begin{bmatrix} s_1 & a_1 \\ s_2 & a_2 \\ \vdots & \vdots \\ s_N & a_N \end{bmatrix}
%     \mathbf{z}^{(0)} &= \mathbf{E}(\mathbf{x}) \quad \text{其中} \quad \mathbf{z}^{(0)} \in \mathbb{R}^{N \times d_{\text{model}}} \\
%     \mathbf{z}^{(0)}_{\text{pos}} &= \text{PositionalEncoding}(\mathbf{z}^{(0)}) \\
%     \mathbf{z}^{(l)} &= \text{EncoderLayer}_l(\mathbf{z}^{(l-1)}) \quad \forall l \in \{1, \dots, M\} \\
%     \mathbf{B}_{i,j} &= \text{Bias}(i, j) = \text{decay\_factor} \times (j - i) \\
%     \mathbf{z}_N &= \mathbf{z}^{(M)}_N \quad \text{其中} \quad \mathbf{z}_N \in \mathbb{R}^{d_{\text{model}}} \\
%     \mathbf{h}_1 &= \text{ReLU}(W_1 \mathbf{z}_N + b_1) \\
%     \mathbf{h}_2 &= \text{ReLU}(W_2 \mathbf{h}_1 + b_2) \\
% \end{align}

% \begin{figure*}[ht]
%     \centering
%     \includegraphics[width=0.5\textwidth, height=8cm, keepaspectratio]{attention view.pdf}
%     \caption{Biased Attention Visualization}
%     \label{fig:attention_visualization}
% \end{figure*}

\subsection{Hierarchical Experience Replay Mechanism}
\label{Hierarchical Experience Replay Mechanism}

The Transformer-based risk predictor relies on data generated from interaction for its training. However, during the initial training phase, actions are largely random, leading to a high frequency of collision experiences. This can cause the model to converge to an overly conservative policy.

To address this issue, we design a \textbf{hierarchical experience replay mechanism}. The core idea is to store data in separate buffers based on whether an episode concludes with a collision. This results in two distinct buffers:
\begin{itemize}
    \item \textbf{High-Risk Experience Buffer}: Stores trajectory sequences that end in a collision.
    \item \textbf{Standard Experience Buffer}: Stores other routine, safe trajectory sequences where the vehicle reaches its destination.
\end{itemize}

During the agent's interaction with the environment, when a complete trajectory concludes, it is labeled accordingly: if the episode ends with a collision, the trajectory sequence is assigned a label of 1; if the vehicle successfully reaches its destination, it receives a label of 0. The complete, labeled sequence is then stored in the corresponding buffer. Trajectories shorter than the required sequence length $N$ are padded to ensure uniform input size.

For model training, we employ a \textbf{balanced sampling strategy} in place of traditional random sampling. Specifically, a batch of size $B$ is composed of two sub-batches of size $B/2$, which are sampled independently from the high-risk and standard experience buffers, respectively. This ensures that the model is trained on an equal proportion of safe and unsafe outcomes, preventing the learning process from being dominated by one type of experience. The overall process is summarized in Algorithm~\ref{alg:main}.

\section{EXPERIMENTS}
\label{EXPERIMENTS}
To validate the effectiveness of the proposed decision framework, this study establishes a simulation environment based on the SUMO platform. This section first introduces the experimental setup and evaluation metrics. Subsequently, it presents the baseline algorithms. The model is then analyzed from the two dimensions of safety and efficiency. Finally, the effectiveness of the risk monitoring model is verified through visualization.
\subsection{Experimental Setup}

To validate the effectiveness of the decision-making framework proposed in this paper, we built a high-fidelity simulation environment based on the open-source microscopic traffic simulation platform SUMO (Simulation of Urban Mobility). To enhance the authenticity of the simulation model, the traffic flow data used in the experiments (including turning traffic at each intersection) is sourced from the real-world dataset of Colorado Springs, USA.\cite{10wang2025learning} Real-time interaction between the simulation platform and Python control algorithms is achieved through the TraCI interface. The experimental scenario is a symmetric four-way intersection, with each direction having one shared lane for straight-through and left-turn movements and a dedicated right-turn lane. The speed limit for all roads in the scenario is set to 10 m/s, and the decision control area for the agent is defined as the 30-meter range before entering the intersection.

% To ensure fairness in comparative experiments, all models based on the Soft Actor-Critic (SAC) algorithm use the same parameter configuration. Specifically, the Actor and Critic networks of the SAC model both use a multi-layer perceptron (MLP) with two hidden layers, each containing 256 nodes. The core hyperparameters are as follows: the learning rate for the Actor network is 0.0003, the learning rate for the Critic network is 0.0004, the batch size is 256, and the experience replay pool capacity is set to 1,000,000. The initial entropy coefficient is set to 0.12, the activation function is ReLU, and each model is trained for 500 episodes.

% For the Risk Predictor, the embedding layer dimension is set to 128, and the Transformer module consists of 2 encoder layers with 4 attention heads. The learning rate for this model is 1e-5, the batch size is 128, and the binary cross-entropy (BCELoss) is used as the loss function.

To ensure a fair comparison in the experiments, all models based on the Soft Actor-Critic (SAC) algorithm adopt a consistent set of hyperparameters. Specifically, the Actor and Critic networks in the SAC model both utilize multi-layer perceptrons (MLP) with two hidden layers, each containing 256 nodes. The core hyperparameters are as follows: the learning rate for the Actor Network is 0.0003, the learning rate for the Critic Network is 0.0004, the batch size is 256, and the experience replay pool capacity is 1,000,000. The initial value of the entropy coefficient is set to 0.12, and the activation function is uniformly set to ReLU. Each model is trained for 500 episodes, with a step length of 1000. The rewards are set as $\lambda_{\text{eff}}$ is 1, $\lambda_{\text{risk}}$ is 3, and $\lambda_{\text{safe}}$ is 10.

For the Risk Predictor, the embedding layer dimension is set to 128, the Transformer module consists of 2 encoder layers, and utilizes 4 attention heads. The learning rate for this model is 1e-5, with a batch size of 128, and $\beta$ is set to 0.2. The sequence length is 10, the size of the risk experience pool and the safe experience pool is 10,000, and binary cross-entropy (BCELoss) is used as the loss function.

All experiments were trained on a device equipped with an NVIDIA RTX-4060Ti (16GB VRAM).

\subsection{Evaluation Metrics}

To intuitively evaluate the performance of the proposed framework, we selected the following three core evaluation metrics from the dimensions of traffic efficiency and safety:

\paragraph{Average Waiting Time(AWT)}
Defined as the average time vehicles spend waiting within the intersection area. This metric is crucial for measuring traffic efficiency, directly reflecting the congestion and delay levels at the intersection. A lower AWT indicates higher traffic efficiency.

\paragraph{Average Queue Length(AQL)}
Refers to the average number of vehicles queuing on each entrance lane of the intersection. This metric not only reflects the severity of traffic congestion but also indirectly measures the utilization efficiency of road capacity. A shorter AQL implies lighter congestion and more efficient utilization of road resources.

\paragraph{Collision Rate(CR)}
As the core metric for safety assessment, this is defined as the ratio of total collision incidents to the total traffic volume successfully passing through the intersection within a specific evaluation period. A lower CR directly corresponds to a higher level of traffic safety.

\begin{algorithm}
\SetAlgoLined
\DontPrintSemicolon
\caption{SAC with Risk Prediction}
\label{alg:main}
Initialize Actor network $\pi_{\phi}$, Critic networks $Q_{\theta_1}, Q_{\theta_2}$;

Initialize Target Critic networks with parameters $\theta'_1 \leftarrow \theta_1, \theta'_2 \leftarrow \theta_2$\;

Initialize Transformer Risk Predictor $P_{\psi}$

Initialize Hierarchical Experience Replay Buffers:$\mathcal{D}_{\text{SAC}}$,   $\mathcal{D}_{\text{safe}}$ and $\mathcal{D}_{\text{risk}}$\;

\For{each episode}{
    Reset environment and receive initial state $s_1$\;
    
    \For{each step $t=1, \dots, T$}{
        Select action $a_t \sim \pi_{\phi}(\cdot|s_t)$\;
        
        Execute action $a_t$, get next state $s_{t+1}$ and efficiency reward $r_{\text{eff}}$\;

        Get current state-action sequence $X_t$ of length $S$ from $\mathcal{B}_{\text{traj}}$\;

        Predict risk: $r_\text{Risk} = P_{\psi}(X_t)$\;
        
        Calculate final reward: $r_t$;
        
        Store current transition $(s_t, a_t)$ in temporary buffer $\mathcal{B}_{\text{traj}}$\;

        Store complete transition $(s_t, a_t, r_t, s_{t+1})$ in a temporary buffer $\mathcal{D}_{\text{SAC}}$\;
        
        \If{vehicle ends due to a collision}{
            Label trajectory sequence $X_t$ with \texttt{label=1}\;
            Store labeled sequence in $\mathcal{D}_{\text{risk}}$\;
        }
        \ElseIf{vehicle reaches destination}{
            Label trajectory sequence $X_t$ with \texttt{label=0}\;
            Store labeled sequence in $\mathcal{D}_{\text{safe}}$\;
        }
    }
    Take a batch of samples from $\mathcal{D}_{\text{SAC}}$ and updatae the critic network and actor network;
    
    Train Risk Predictor $P_{\psi}$ using labeled data from $\mathcal{D}_{\text{safe}}$ and $\mathcal{D}_{\text{risk}}$ with BCE loss\;
}

\end{algorithm}

\begin{figure}[!t]
    \centering
    
    % 第一个子图
    \subfloat[]{\includegraphics[width=3.2in]{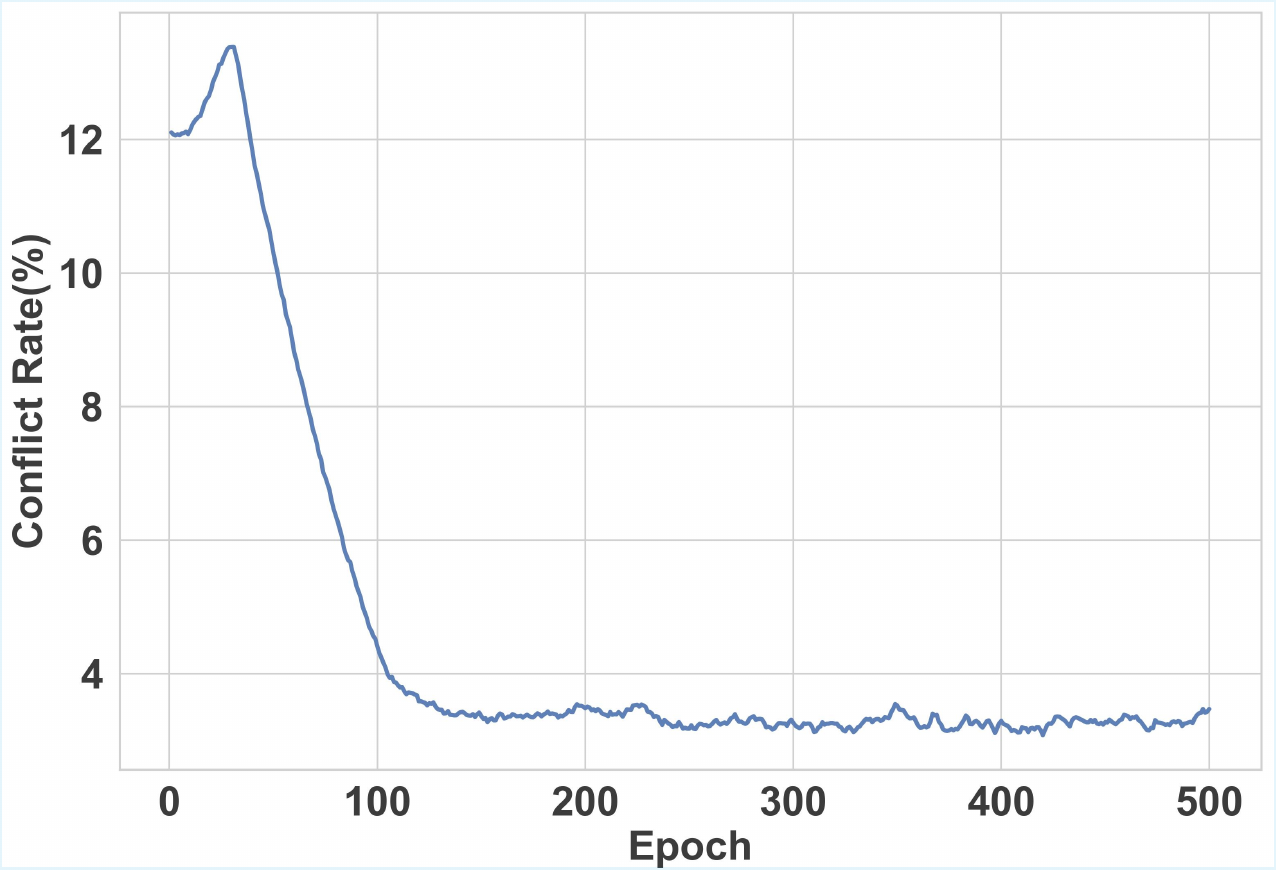}
    \label{fig:conflict_shou}
    }
    
    % 第二个子图
    \subfloat[]{\includegraphics[width=3.2in]{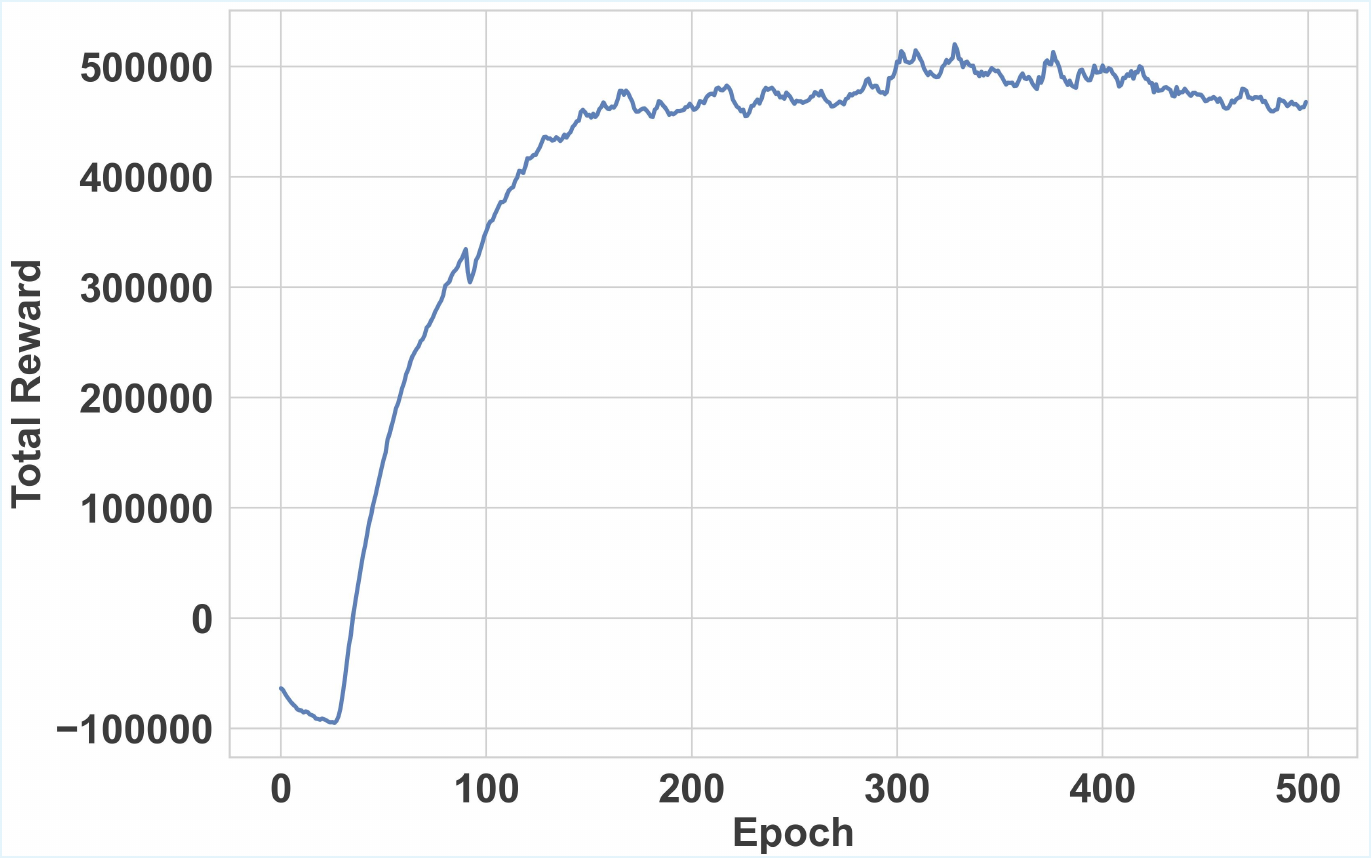}
    \label{fig:reward_shou}
    }
    
    \caption{Convergence Analysis of the Model Training Process (a) Conflict Rate Convergence. (b) Average Reward Convergence. \textbf{Note:} Vehicles with speed less than 0.1 m/s are considered to have collided.}
    \label{fig:shoulian}

\end{figure}

\subsection{Comparative experiment}
To comprehensively evaluate the performance of our proposed framework, we selected two representative methods as baselines, encompassing approaches from traditional rules to advanced reinforcement learning algorithms.

First-Come, First-Served (FCFS) \cite{15dresner2008multiagent}: This is a classic rule-based coordination strategy for unsignalized intersections. Under this strategy, all vehicles entering the control area (RVs) are assigned a timestamp upon arrival. The right-of-way is granted to the vehicle that first arrives at the intersection stop line. In the event that multiple vehicles arrive simultaneously, one is randomly selected to be granted the right of way.

Dawei Wang et al. \cite{10wang2025learning}: This represents one of the current state-of-the-art (SOTA) reinforcement learning methods for mixed traffic control at complex real-world intersections. This method also employs a decentralized decision-making framework, where each Robotic Vehicle (RV) makes high-level decisions to “Go” or “Stop” based on observations of the surrounding traffic environment. Furthermore, it integrates a rule-based conflict resolution mechanism: when multiple RVs from conflicting traffic flows all decide to “Go,” only the one with the highest priority (calculated based on waiting time and queue length) is allowed to enter.

As shown in Figure~\ref{fig:shoulian}, the SAC model, after training, enables vehicles to achieve the maximum possible cumulative reward. The reward function takes into account safety, traffic efficiency and global information, leading to a more orderly vehicle movement. Figures~\ref{fig:conflict_shou} and Figure~\ref{fig:reward_shou} illustrate the convergence results for the total reward and the collision rate, respectively. 
It should be noted that IDM incorporates proactive yielding behaviors, which can cause vehicles to brake abruptly and wait inside an intersection. Although such a scenario does not constitute a collision in SUMO, it signifies a critical conflict situation. To address this, we introduce a specific definition for our training phase: a vehicle is considered to have had a “collision” if its speed falls below 0.1 m/s while inside the intersection. Following this event, the vehicle is immediately removed from the simulation. (This particular setting is deactivated for the testing phase). This training strategy enables the collision rate to converge to 3\% \%, which demonstrably reduces the potential for conflict risks.

Table \ref{tab:comparision of different methods} presents a quantitative comparison of the proposed model with the baseline models on three main evaluation metrics, based on the results of 1500 test rounds following the simulation.

Advantages of Reinforcement Learning Methods: Compared to the traditional FCFS model, both reinforcement learning-based methods demonstrate significant advantages in traffic efficiency. The proposed model reduces the average waiting time from 199.26 seconds to 50.81 seconds, a substantial decrease of 74.5\%, while also significantly reducing the average queue length. This demonstrates the immense potential of reinforcement learning in optimizing the utilization of intersection resources.

Superiority of the Proposed Method: Compared to the state-of-the-art Dawei Wang \cite{10wang2025learning} baseline, the proposed “Risk with bias” model achieves superior performance across all metrics. It further reduces the average waiting time and average queue length by 17.5\% and 28.4\%, respectively. Concurrently, the collision rate decreases from 0.19\% to 0.1\%, indicating an enhanced level of safety. These results fully illustrate that the continuous control strategy guided by risk prediction, as proposed in this paper, is more efficient and safer than strategies based on discrete actions and external conflict resolution.

\begin{table}[H]
    \centering % 表格居中
    \caption{Comparison of different methods} % 表格标题
    \label{tab:comparision of different methods} % 用于交叉引用
    \begin{tabular}{lccc}
        \hline
        \textbf{Methods} & \textbf{AQL(veh/s)} & \textbf{AWT(s)} & \textbf{CT(\%)} \\
        \hline
        FCFS\cite{15dresner2008multiagent} & 8.52 & 199.26 & 0 \\
        Wang\cite{10wang2025learning} & 6.41 & 61.57 & 0.19 \\
        SAC-RWB(ours) & 4.59 & 50.81 & 0.1 \\
        \hline
    \end{tabular}
\end{table}

\begin{figure}[!t]
    \centering
    
    % 第一个子图
    \subfloat[]{\includegraphics[width=3.2in]{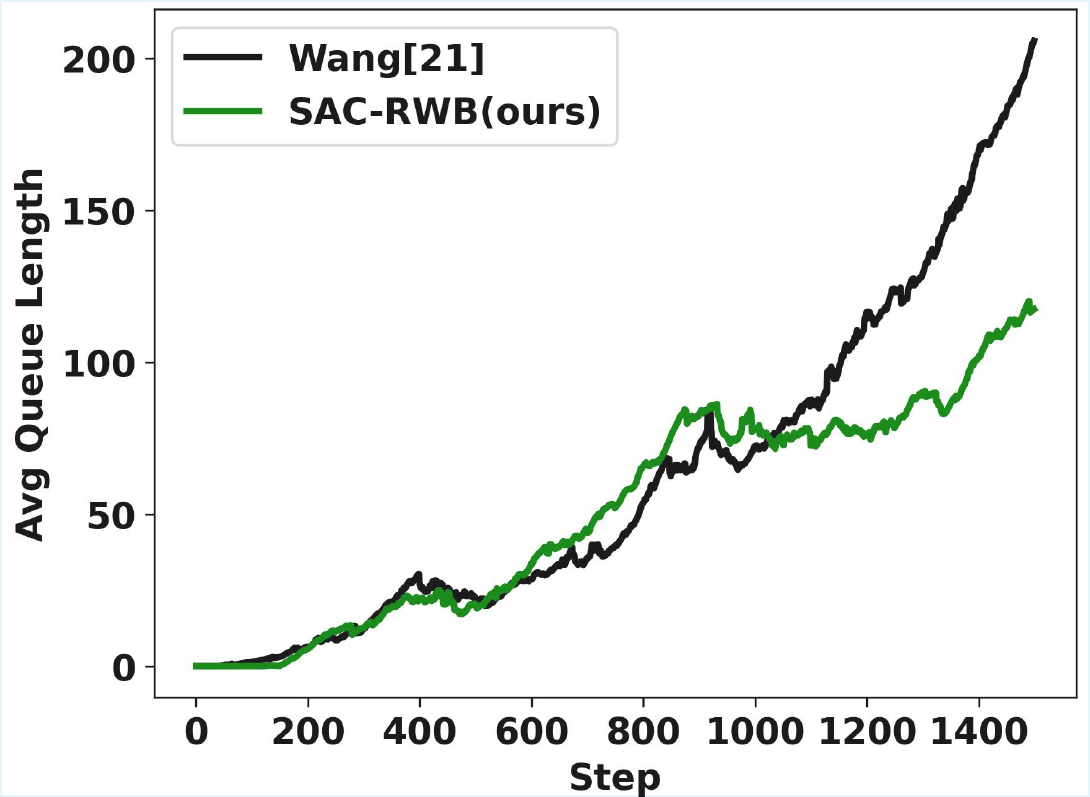}
    \label{fig:comparison_awt_6}
    }
    
    % 第二个子图
    \subfloat[]{\includegraphics[width=3.2in]{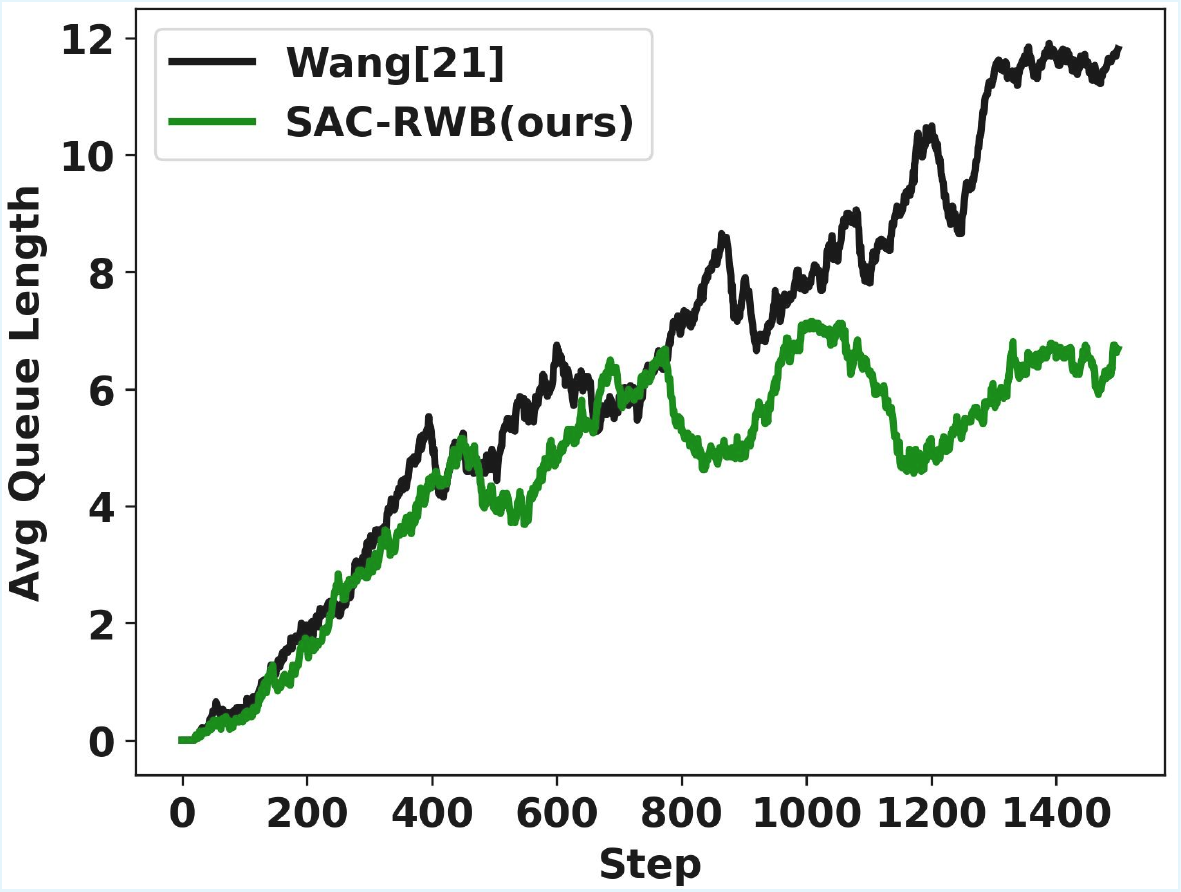}
    \label{fig:comparison_AQL_6}
    }
    
    \caption{Dynamic Performance Comparison against the Baseline Model. (a) Comparison of Average Waiting Time. (b) Comparison of Average Queue Length.}
    \label{fig:Dynamic Performance Comparison}
    
\end{figure}

To further evaluate the dynamic performance of the model under continuous traffic flow, Figure~\ref{fig:Dynamic Performance Comparison} illustrates the trends of various efficiency metrics over simulation time for the proposed model and the Dawei Wang \cite{10wang2025learning} baseline. As shown in Figure~\ref{fig:Dynamic Performance Comparison}, as the number of simulation steps increases (representing the continuous accumulation of traffic volume), the proposed SAC model consistently maintains lower average waiting times and average queue lengths compared to the baseline model. This indicates that our decision framework not only outperforms in overall performance but also exhibits greater stability in response to sustained traffic pressure, effectively preventing the rapid accumulation of congestion.

\subsection{Ablation experiment}
% This section presents an ablation study to analyze the model. First, Figure~\ref{fig:total_reward} and Figures~\ref{fig:efficiency} show the comprehensive reward and efficiency reward, respectively. Since the risk reward and comprehensive reward differ in the “no risk” experiment, only the efficiency rewards of the three configurations are compared, whereas the comprehensive rewards are compared between the Original Attention and Biased Attention models. As shown in Figure~\ref{fig:total_reward}, as training progresses, the reward gradually increases and converges. Moreover, with the introduction of the bias matrix, the overall reward of the model improves to some extent. Figure~\ref{fig:efficiency} compares the efficiency rewards of the three configurations. In the 'no risk' experiment, the efficiency reward of the model remains around 0, indicating lower vehicle travel efficiency. In contrast, after incorporating the risk reward, the efficiency reward significantly improves, effectively improving vehicle travel efficiency.
\begin{figure}[!t]
    \centering    
    \includegraphics[width=3.2in]{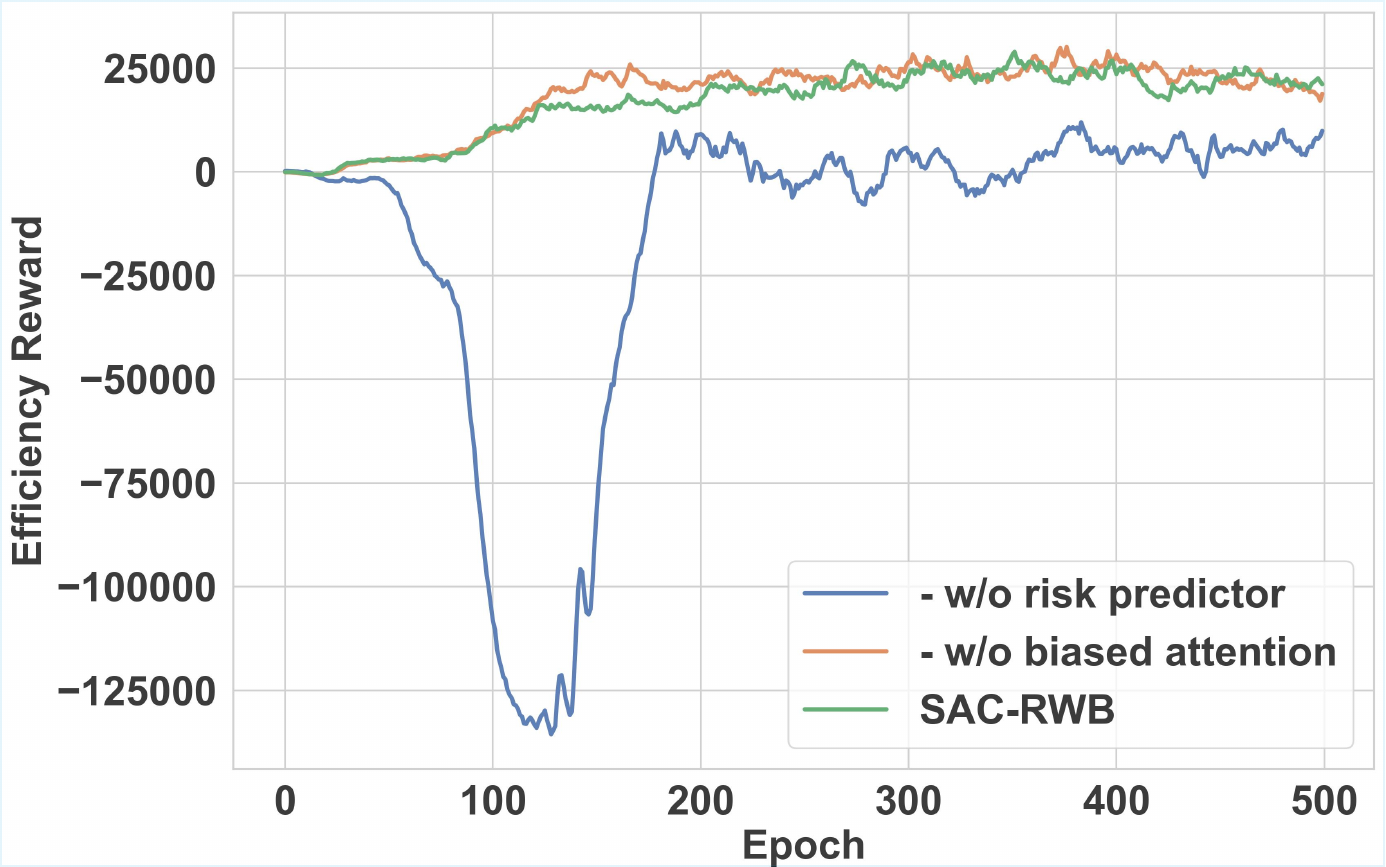}
    \caption{Efficiency reward of the training progress.}
    \label{fig:efficiency7}
\end{figure}

In this section, we conduct an ablation study on our model. First, we define a baseline experiment, termed ``- w/o risk predictor" (without risk predictor), where the risk reward is directly removed from the model's training process. Furthermore, we introduce an experiment, ``- w/o biased attention" (risk predictor without biased attention), which utilizes the risk reward provided by the risk prediction module with the original attention mechanism. Since the total reward is made up of contributions from different modules, we specifically compare the efficiency reward component. As shown in Figure~\ref{fig:efficiency7}, which illustrates the convergence results of the efficiency reward of the model, the results indicate that introducing the risk reward leads to an improvement in the efficiency reward of the model. Moreover, with the introduction of biased attention, the efficiency reward is further enhanced.

\begin{figure}[!t]
    \centering
    
    % 第一个子图
    \subfloat[]{\includegraphics[width=3.2in]{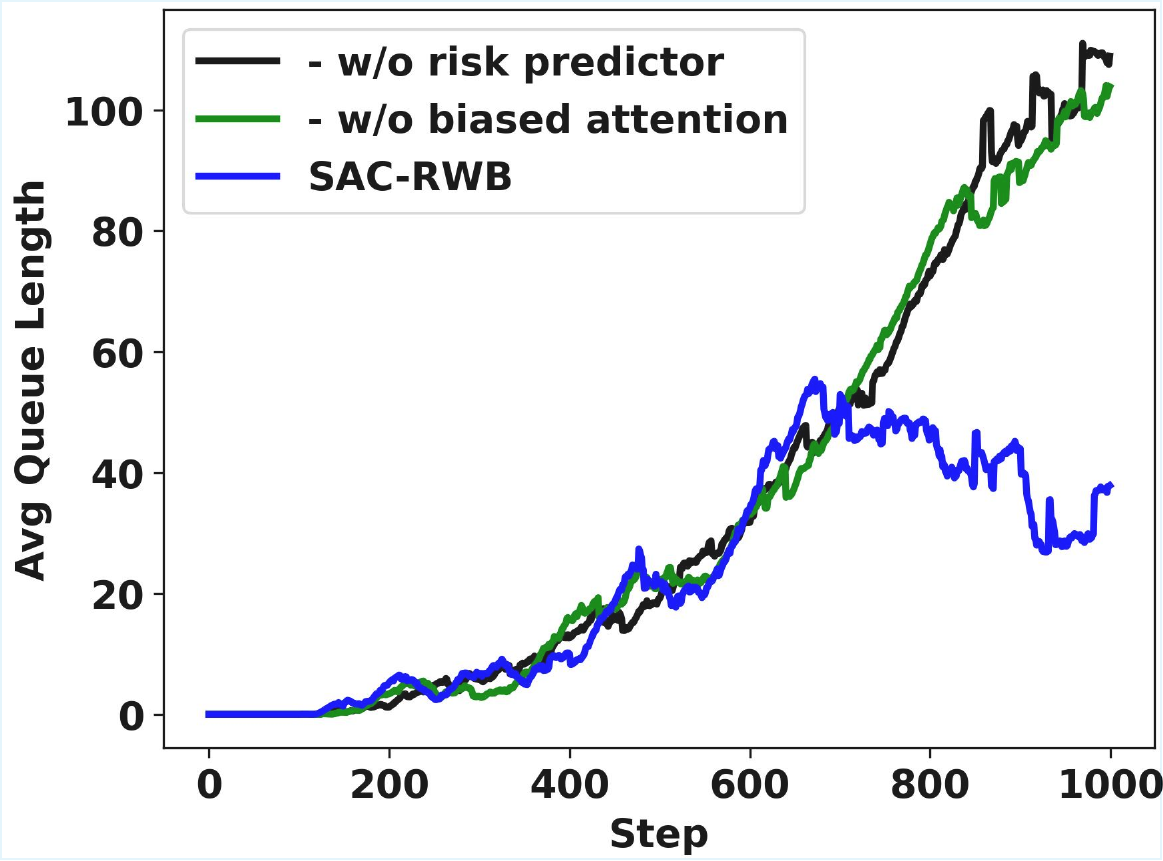}
    \label{fig:waiting_time}
    }%

    % 第二个子图
    \subfloat[]{\includegraphics[width=3.2in]{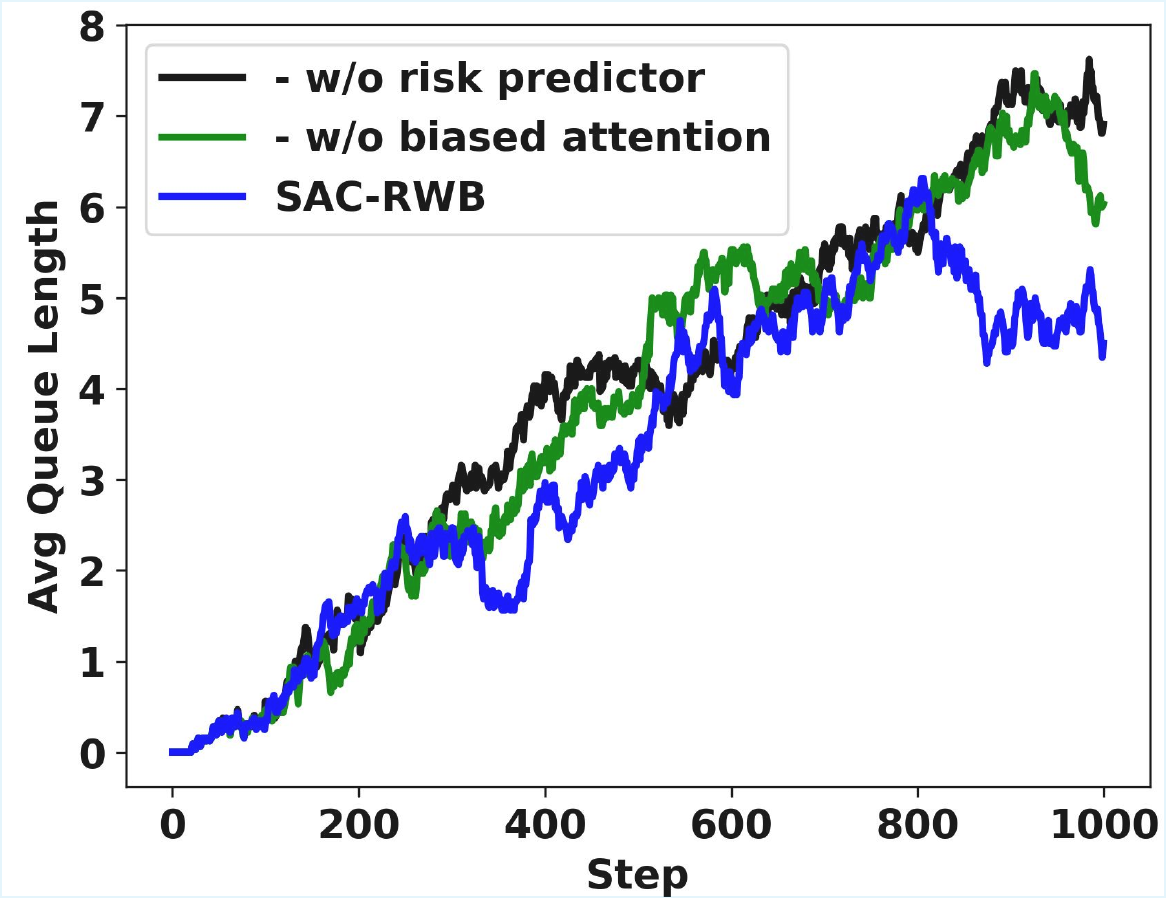}
    \label{fig:queue_length}
    }%

    % 第三个子图
    \subfloat[]{\includegraphics[width=3.2in]{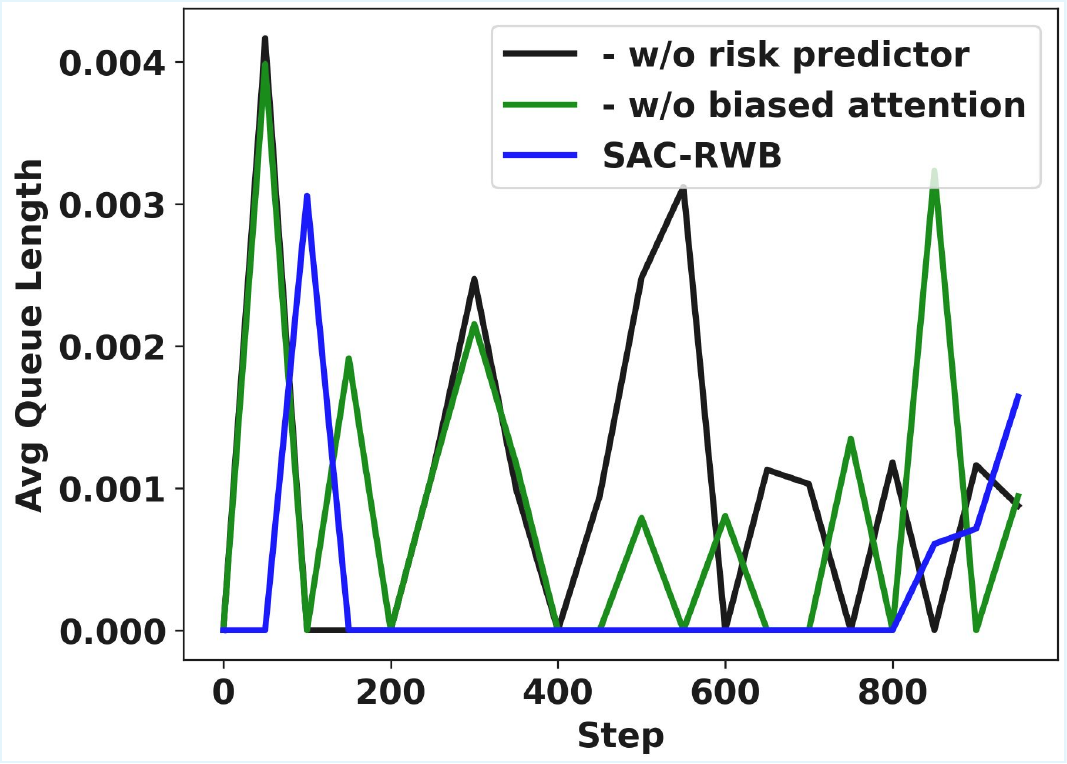}
    \label{fig:conflict_rate}
    }%

    \caption{Performance Comparison and Analysis of Different Models. (a) Comparison of Average Waiting Time. (b) Comparison of Average Queue Length.(c) Comparison of Average Conflict Rate}
    \label{fig:comparison}
\end{figure}

Table \ref{tab:3} presents the comparison results of different models in terms of average waiting time, average queue length, and collision rate. These results were obtained by setting three different seed sets in the experimental scenario, followed by testing the trained models over a simulation time of 1000 steps, after which the results were averaged. After introducing the biased attention-based risk prediction module, further reductions in average waiting time, average queue length, and vehicle collision rate were observed.

\begin{table}[H]
    \centering % 表格居中
    \caption{Comparison of evaluation metrics under different models} % 表格标题
    \label{tab:3} % 用于交叉引用
    \begin{tabular}{lccc}
        \hline
        \textbf{Methods} & \textbf{AQL(veh)} & \textbf{AWT(s)} & \textbf{CT(\%)} \\
        \hline
        - w/o risk predictor & 31.67 & 3.71 & 0.068 \\
        - w/o biased attention & 28.37 & 3.67 & 0.076 \\
        SAC-RWB & 21.08 & 3.24 & 0.061 \\
        \hline
    \end{tabular}
\end{table}

\begin{figure}[t]
    \centering
    \includegraphics[width=0.48\textwidth]{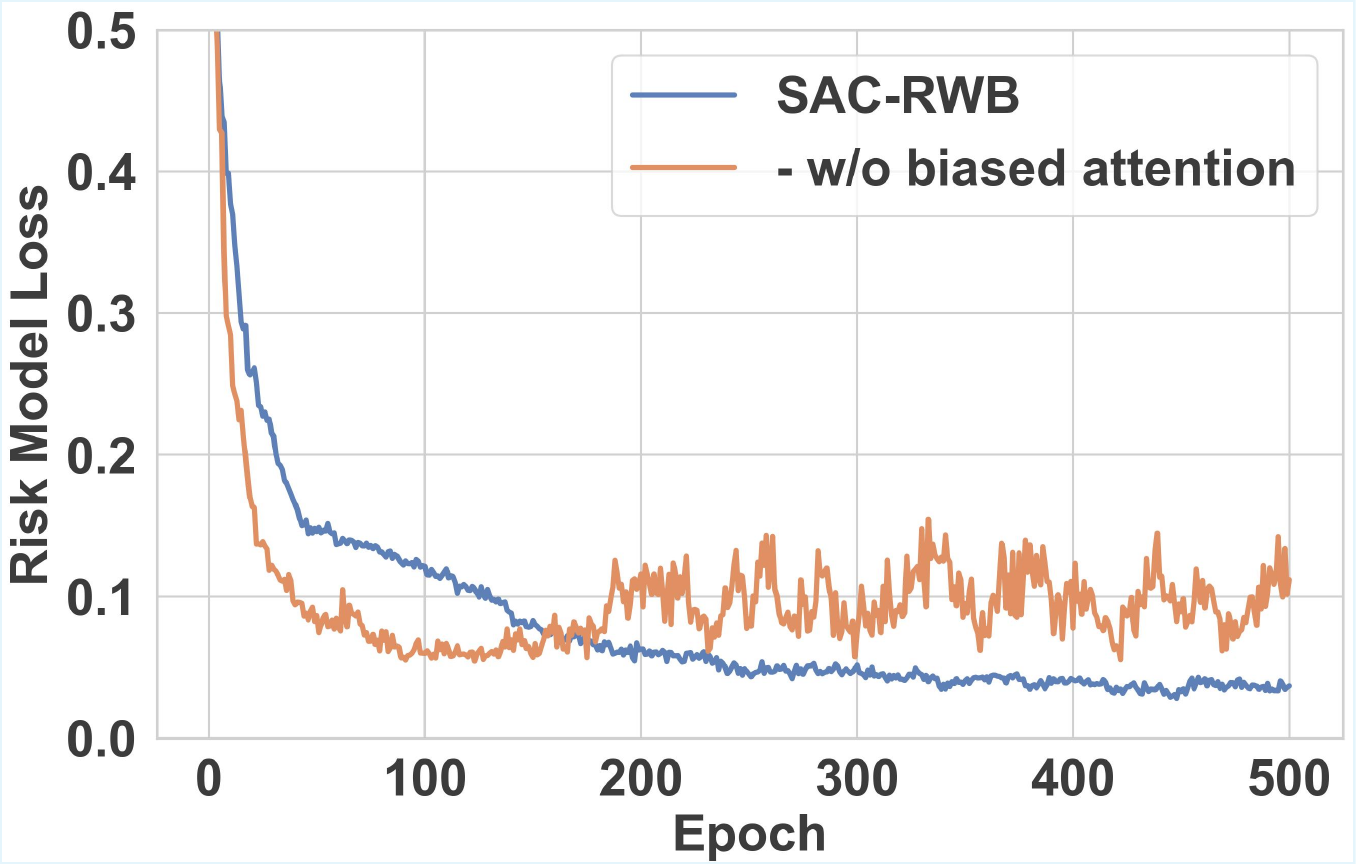}
    \caption{Risk model loss of the progress}
    \label{fig:risk_loss} % 设置标签，用于正文引用
\end{figure}

\begin{figure*}[!t]
    \centering
    
    % 第一个子图
    \subfloat[]{\includegraphics[width=0.48\textwidth]{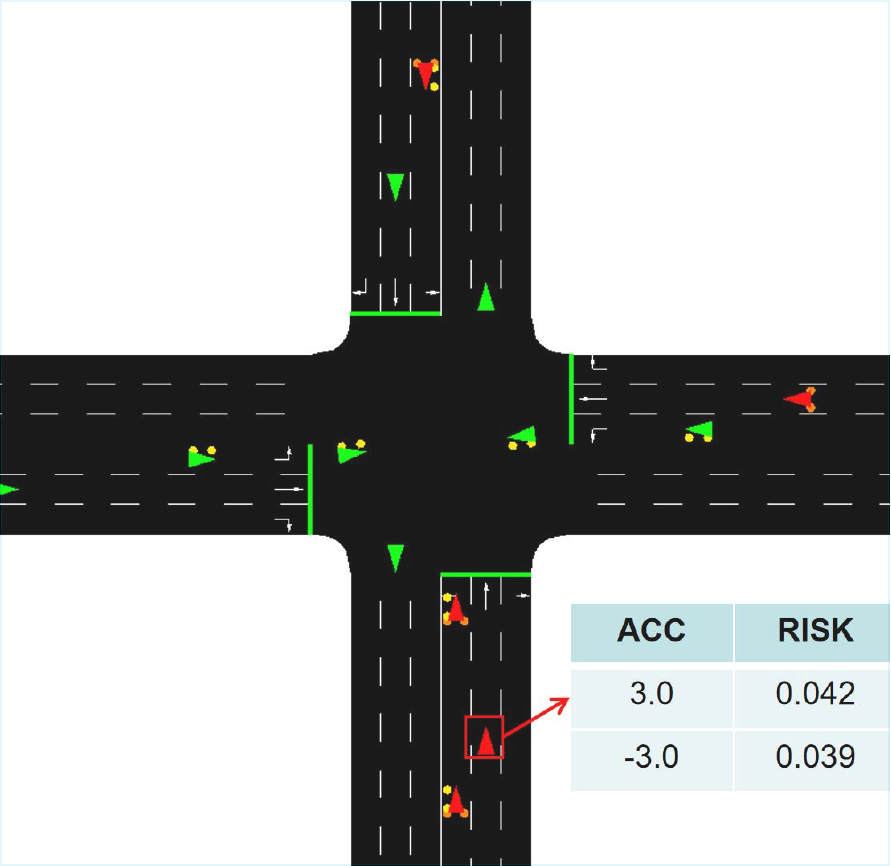}
    \label{fig:original_analysis}
    }%
    \hfill
    % 第二个子图
    \subfloat[]{\includegraphics[width=0.48\textwidth]{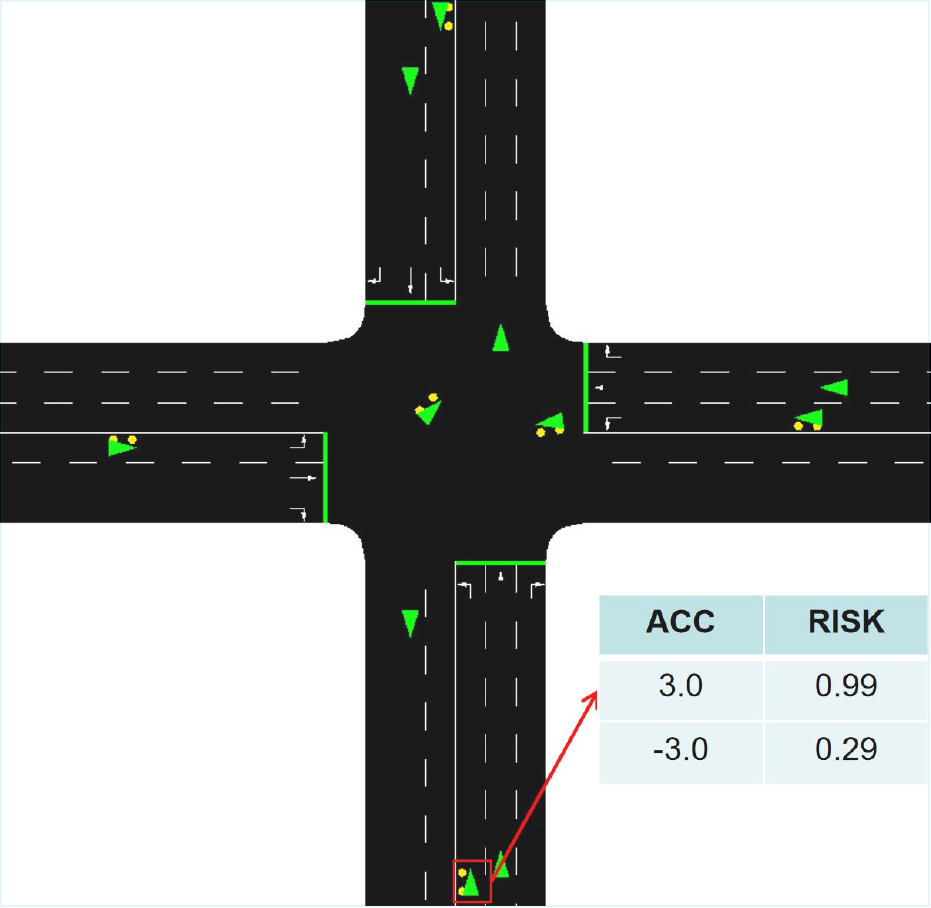}
    \label{fig:Biased_analysis}
    }%
    
    \caption{Action Sensitivity Analysis of the Risk Prediction Model.(a) Action Sensitivity Analysis of the Original Attention Model. (b) Action Sensitivity Analysis of the Biased Attention Model.}
    \label{fig:risk_analysis}
\end{figure*}

Figure~\ref{fig:comparison} illustrates the test results under different models. Figure~\ref{fig:waiting_time}, Figure~\ref{fig:queue_length}, and Figure~\ref{fig:conflict_rate} depict the variation curves of average waiting time, average queue length, and average collision rate, respectively. As shown in Figure~\ref{fig:waiting_time} and Figure~\ref{fig:queue_length}, with the increase in traffic flow, the proposed method, after the introduction of the risk predictor, can still effectively control traffic flow while maintaining a low collision rate, demonstrating high driving safety.

\subsection{Analysis of the Risk Prediction Model}

To validate the superiority of the proposed biased attention mechanism, we conducted an ablation study comparing it with a model employing standard self-attention.

First, regarding the model training process, the loss convergence curves for both attention mechanisms are illustrated in Figure~\ref{fig:risk_loss}. As clearly observed, the introduction of biased attention results in a more stable decline in loss and convergence to a lower final value. This indicates that the biased attention mechanism facilitates more efficient learning of the risk assessment task.

To intuitively demonstrate the differences in decision sensitivity between the two mechanisms, we designed a specific scenario for testing. As depicted in Figure~\ref{fig:original_analysis} and Figure~\ref{fig:Biased_analysis}, we selected a traffic state with potential collision risk and manually modified the last action (acceleration) of the current vehicle to observe changes in the predicted risk value.

\begin{itemize}
    \item \textbf{Standard Attention Model}: As shown in Figure~\ref{fig:original_analysis}, when using the standard attention model, even switching the vehicle’s acceleration from 3.0 m/s² (acceleration) to -3.0 m/s² (deceleration) resulted in only a marginal decrease in the predicted risk value, from 0.042 to 0.039. This negligible change suggests that the model’s attention failed to focus on the recent actions most critical for risk judgment, leading to insensitivity to changes in current decisions.
    \item \textbf{Biased Attention Model}: In contrast, as presented in Figure~\ref{fig:Biased_analysis}, the model with biased attention exhibited exceptional decision sensitivity. Under identical conditions, when the action was acceleration (3.0 m/s²), the model predicted a high risk of 0.99; upon switching to deceleration (-3.0 m/s²), the predicted risk plummeted to 0.29.
\end{itemize}

In summary, this comparative experiment clearly demonstrates that the biased attention mechanism enables the model to more effectively focus on recent state-action pairs, allowing it to associate current decisions with future risks accurately. Such heightened sensitivity to immediate actions is crucial for reinforcement learning agents to learn proactive safe driving strategies.

\section{Conclusion}
\label{Conclusion}
To address the complex decision-making challenges at unsignalized intersections, this paper proposes a Deep Reinforcement Learning (DRL) framework integrated with biased attention. This framework is built upon the Soft Actor-Critic (SAC) algorithm, with its core being a risk predictor based on biased attention. This predictor transforms long-term collision risks into dense reward signals, guiding the agent to achieve proactive and safe decisions. Combined with a hierarchical experience replay mechanism, the framework effectively accelerates the model’s convergence speed. Simulation results demonstrate that, compared to baseline models, the proposed framework exhibits significant advantages in reducing the collision rate and shortening waiting times.

Future research will extend the controlled area from the intersection entrance to the interior of the intersection. The current model primarily focuses on access control at the entrance. By incorporating the intersection’s interior into refined management, real-time planning, and dynamic coordination of vehicle trajectories through conflict areas can be achieved. This would further mitigate potential collision risks and enhance the overall system safety.


\begin{thebibliography}{10}
\providecommand{\url}[1]{#1}
\csname url@samestyle\endcsname
\providecommand{\newblock}{\relax}
\providecommand{\bibinfo}[2]{#2}
\providecommand{\BIBentrySTDinterwordspacing}{\spaceskip=0pt\relax}
\providecommand{\BIBentryALTinterwordstretchfactor}{4}
\providecommand{\BIBentryALTinterwordspacing}{\spaceskip=\fontdimen2\font plus
\BIBentryALTinterwordstretchfactor\fontdimen3\font minus \fontdimen4\font\relax}
\providecommand{\BIBforeignlanguage}[2]{{%
\expandafter\ifx\csname l@#1\endcsname\relax
\typeout{** WARNING: IEEEtran.bst: No hyphenation pattern has been}%
\typeout{** loaded for the language `#1'. Using the pattern for}%
\typeout{** the default language instead.}%
\else
\language=\csname l@#1\endcsname
\fi
#2}}
\providecommand{\BIBdecl}{\relax}
\BIBdecl

\bibitem{1}
A.~Haydari and Y.~Y{\i}lmaz, ``Deep reinforcement learning for intelligent transportation systems: A survey,'' \emph{IEEE Transactions on Intelligent Transportation Systems}, vol.~23, no.~1, pp. 11--32, 2020.

\bibitem{2}
D.~Fajardo, T.-C. Au, S.~T. Waller, P.~Stone, and D.~Yang, ``Automated intersection control: Performance of future innovation versus current traffic signal control,'' \emph{Transportation Research Record}, vol. 2259, no.~1, pp. 223--232, 2011.

\bibitem{3}
D.~Isele and A.~Cosgun, ``Transferring autonomous driving knowledge on simulated and real intersections,'' \emph{arXiv preprint arXiv:1712.01106}, 2017.

\bibitem{4}
C.~Chen, B.~Wu, L.~Xuan, J.~Chen, and L.~Qian, ``A discrete control method for the unsignalized intersection based on cooperative grouping,'' \emph{IEEE Transactions on Vehicular Technology}, vol.~71, no.~1, pp. 123--136, 2021.

\bibitem{5}
X.~Gong, B.~Wang, and S.~Liang, ``Collision-free cooperative motion planning and decision-making for connected and automated vehicles at unsignalized intersections,'' \emph{IEEE Transactions on Systems, Man, and Cybernetics: Systems}, vol.~54, no.~5, pp. 2744--2756, 2024.

\bibitem{6bouton2019reinforcement}
M.~Bouton, J.~Karlsson, A.~Nakhaei, K.~Fujimura, M.~J. Kochenderfer, and J.~Tumova, ``Reinforcement learning with probabilistic guarantees for autonomous driving,'' \emph{arXiv preprint arXiv:1904.07189}, 2019.

\bibitem{7yang2024towards}
K.~Yang, S.~Li, Y.~Chen, D.~Cao, and X.~Tang, ``Towards safe decision-making for autonomous vehicles at unsignalized intersections,'' \emph{IEEE Transactions on Vehicular Technology}, 2024.

\bibitem{8spatharis2024multiagent}
C.~Spatharis and K.~Blekas, ``Multiagent reinforcement learning for autonomous driving in traffic zones with unsignalized intersections,'' \emph{Journal of Intelligent Transportation Systems}, vol.~28, no.~1, pp. 103--119, 2024.

\bibitem{9xiao2024decision}
W.~Xiao, Y.~Yang, X.~Mu, Y.~Xie, X.~Tang, D.~Cao, and T.~Liu, ``Decision-making for autonomous vehicles in random task scenarios at unsignalized intersection using deep reinforcement learning,'' \emph{IEEE Transactions on Vehicular Technology}, vol.~73, no.~6, pp. 7812--7825, 2024.

\bibitem{10wang2025learning}
D.~Wang, W.~Li, L.~Zhu, and J.~Pan, ``Learning to control and coordinate mixed traffic through robot vehicles at complex and unsignalized intersections,'' \emph{The International Journal of Robotics Research}, vol.~44, no.~5, pp. 805--825, 2025.

\bibitem{11yan2021courteous}
S.~Yan, T.~Welschehold, D.~B{\"u}scher, and W.~Burgard, ``Courteous behavior of automated vehicles at unsignalized intersections via reinforcement learning,'' \emph{IEEE Robotics and Automation Letters}, vol.~7, no.~1, pp. 191--198, 2021.

\bibitem{12treiber2000congested}
M.~Treiber, A.~Hennecke, and D.~Helbing, ``Congested traffic states in empirical observations and microscopic simulations,'' \emph{Physical review E}, vol.~62, no.~2, p. 1805, 2000.

\bibitem{13haarnoja2018soft}
T.~Haarnoja, A.~Zhou, K.~Hartikainen, G.~Tucker, S.~Ha, J.~Tan, V.~Kumar, H.~Zhu, A.~Gupta, P.~Abbeel \emph{et~al.}, ``Soft actor-critic algorithms and applications,'' \emph{arXiv preprint arXiv:1812.05905}, 2018.

\bibitem{14vaswani2017attention}
A.~Vaswani, N.~Shazeer, N.~Parmar, J.~Uszkoreit, L.~Jones, A.~N. Gomez, {\L}.~Kaiser, and I.~Polosukhin, ``Attention is all you need,'' \emph{Advances in neural information processing systems}, vol.~30, 2017.

\bibitem{15dresner2008multiagent}
K.~Dresner and P.~Stone, ``A multiagent approach to autonomous intersection management,'' \emph{Journal of artificial intelligence research}, vol.~31, pp. 591--656, 2008.

\bibitem{16zhang2024car}
T.~T. Zhang, P.~J. Jin, S.~T. McQuade, A.~Bayen, and B.~Piccoli, ``Car-following models: A multidisciplinary review,'' \emph{IEEE Transactions on Intelligent Vehicles}, 2024.

\bibitem{17press2021train}
O.~Press, N.~A. Smith, and M.~Lewis, ``Train short, test long: Attention with linear biases enables input length extrapolation,'' \emph{arXiv preprint arXiv:2108.12409}, 2021.

\bibitem{18li2024decision}
S.~Li, K.~Peng, F.~Hui, Z.~Li, C.~Wei, and W.~Wang, ``A decision-making approach for complex unsignalized intersection by deep reinforcement learning,'' \emph{IEEE Transactions on Vehicular Technology}, vol.~73, no.~11, pp. 16\,134--16\,147, 2024.

\bibitem{19pattanayak2024deep}
S.~K. Pattanayak, M.~Bhoyar, and T.~Adimulam, ``Deep reinforcement learning for complex decision-making tasks,'' \emph{International Journal of Innovative Research in Science, Engineering and Technology}, vol.~13, no.~11, pp. 18\,205--18\,220, 2024.


\bibitem{21mokhtari2021pedestriancollisionavoidanceautonomous}
\BIBentryALTinterwordspacing
K.~Mokhtari and A.~R. Wagner, ``Pedestrian collision avoidance for autonomous vehicles at unsignalized intersection using deep q-network,'' 2021. [Online]. Available: \url{https://arxiv.org/abs/2105.00153}
\BIBentrySTDinterwordspacing

\bibitem{23lu2024activead}
H.~Lu, X.~Jia, Y.~Xie, W.~Liao, X.~Yang, and J.~Yan, ``Activead: Planning-oriented active learning for end-to-end autonomous driving,'' \emph{arXiv preprint arXiv:2403.02877}, 2024.

\bibitem{24chitta}
\BIBentryALTinterwordspacing
K.~Chitta, A.~Prakash, B.~Jaeger, Z.~Yu, K.~Renz, and A.~Geiger, ``Transfuser: Imitation with transformer-based sensor fusion for autonomous driving,'' 2022. [Online]. Available: \url{https://arxiv.org/abs/2205.15997}
\BIBentrySTDinterwordspacing

\bibitem{258845649}
Y.~Chen, J.~Zha, and J.~Wang, ``An autonomous t-intersection driving strategy considering oncoming vehicles based on connected vehicle technology,'' \emph{IEEE/ASME Transactions on Mechatronics}, vol.~24, no.~6, pp. 2779--2790, 2019.

\bibitem{269294407}
C.-J. Hoel, T.~Tram, and J.~Sjöberg, ``Reinforcement learning with uncertainty estimation for tactical decision-making in intersections,'' in \emph{2020 IEEE 23rd International Conference on Intelligent Transportation Systems (ITSC)}, 2020, pp. 1--7.

\bibitem{2710537094}
W.~Li, Y.~Zhang, L.~Li, Y.~Lv, and M.~Wang, ``A pedestrian trajectory prediction model for right-turn unsignalized intersections based on game theory,'' \emph{IEEE Transactions on Intelligent Transportation Systems}, vol.~25, no.~8, pp. 9643--9658, 2024.

\bibitem{29CHEN2022127953}
\BIBentryALTinterwordspacing
X.~Chen, M.~Hu, B.~Xu, Y.~Bian, and H.~Qin, ``Improved reservation-based method with controllable gap strategy for vehicle coordination at non-signalized intersections,'' \emph{Physica A: Statistical Mechanics and its Applications}, vol. 604, p. 127953, 2022. [Online]. Available: \url{https://www.sciencedirect.com/science/article/pii/S0378437122006033}
\BIBentrySTDinterwordspacing

\bibitem{3010.1007/s10489-011-0322-z}
\BIBentryALTinterwordspacing
J.~Wu, A.~Abbas-Turki, and A.~El~Moudni, ``Cooperative driving: an ant colony system for autonomous intersection management,'' \emph{Applied Intelligence}, vol.~37, no.~2, p. 207–222, Sep. 2012. [Online]. Available: \url{https://doi.org/10.1007/s10489-011-0322-z}
\BIBentrySTDinterwordspacing

\bibitem{31AHMANE201344}
\BIBentryALTinterwordspacing
M.~Ahmane, A.~Abbas-Turki, F.~Perronnet, J.~Wu, A.~E. Moudni, J.~Buisson, and R.~Zeo, ``Modeling and controlling an isolated urban intersection based on cooperative vehicles,'' \emph{Transportation Research Part C: Emerging Technologies}, vol.~28, pp. 44--62, 2013, euro Transportation: selected paper from the EWGT Meeting, Padova, September 2009. [Online]. Available: \url{https://www.sciencedirect.com/science/article/pii/S0968090X12001374}
\BIBentrySTDinterwordspacing

\bibitem{32LI2025107960}
\BIBentryALTinterwordspacing
W.~Li, X.~Li, L.~Li, Y.~Tang, and Y.~Hu, ``Simulation of human–vehicle interaction at right-turn unsignalized intersections: A game-theoretic deep maximum entropy inverse reinforcement learning method,'' \emph{Accident Analysis \& Prevention}, vol. 214, p. 107960, 2025. [Online]. Available: \url{https://www.sciencedirect.com/science/article/pii/S0001457525000466}
\BIBentrySTDinterwordspacing

\bibitem{34yu2025uncertainty}
R.~Yu, Z.~Li, L.~Xiong, W.~Han, and B.~Leng, ``Uncertainty-aware safety-critical decision and control for autonomous vehicles at unsignalized intersections,'' \emph{arXiv preprint arXiv:2505.19939}, 2025.

\end{thebibliography}
\end{document}